\documentclass[runningheads]{llncs}
\PassOptionsToPackage{table,dvipsnames}{xcolor}

\usepackage{eccv}

\usepackage{eccvabbrv}

\usepackage{tabularx}
\usepackage{graphicx}
\usepackage{booktabs}

\usepackage{amsfonts,amsmath,amssymb} 
\usepackage{bbold} 
\usepackage{pifont} 

\usepackage{adjustbox} 
\usepackage{makecell} 
\usepackage{multirow} 
\usepackage{comment} 

\usepackage{algorithm, algorithmic} 

\usepackage[accsupp]{axessibility}  % Improves PDF readability for those with disabilities.

\definecolor{mygreen}{RGB}{0,150,0}
\definecolor{myred}{RGB}{190,0,0}
\newcommand{\cmark}{\Large\color{mygreen}\checkmark}
\newcommand{\xmark}{\Large\color{myred}\ding{55}}

\usepackage{hyperref}

\usepackage{orcidlink}

\makeatletter
\renewcommand*{\@fnsymbol}[1]{%
  \ensuremath{\ifcase#1\or \star\or \dagger\or \ddagger\or
  \mathsection\or \mathparagraph\fi}}
\makeatother

\begin{document}

\title{EraseLoRA: MLLM-Driven Foreground Exclusion and Background Subtype Aggregation for Dataset-Free Object Removal} 

\titlerunning{EraseLoRA: Dataset-Free Object Removal}

\author{
Sanghyun Jo\inst{1,2}\thanks{Equal contribution.}\thanks{Corresponding authors: \texttt{shjo.april@gmail.com}, \texttt{kyskim@snu.ac.kr}}
\orcidlink{0000-0001-5371-2069} \and
Donghwan Lee\inst{2}\protect\footnotemark[1]\orcidlink{0009-0006-4719-0042} \and
Eunji Jung\inst{2}\protect\footnotemark[1]\orcidlink{0009-0005-9438-2912} \and
Seong Je Oh\inst{2}\orcidlink{0000-0002-3543-8745} \and
Kyungsu Kim\inst{2}\protect\footnotemark[2]\orcidlink{0000-0001-6622-6545}
}

\authorrunning{S.~Jo et al.}

% Springer Heidelberg, Tiergartenstr.~17, 69121 Heidelberg, Germany
\institute{OGQ, Seoul, Korea \and
Seoul National University, Seoul, Korea}

\maketitle

\begin{abstract}
    Object removal must prevent the masked target from reappearing and reconstruct the occluded background with structural and contextual fidelity, rather than merely filling a hole plausibly. Recent dataset-free approaches manipulate the diffusion model's internal self-attention to prevent it from referencing the masked region, yet they fail in two critical ways: (i) they treat the masked region as the sole foreground, misinterpreting non-target objects as background and regenerating them, and (ii) they apply uniform attention constraints without distinguishing diverse background subtypes, leading to textural blurring and structural misalignment. Both failures stem from the absence of explicit background-aware reasoning. We propose EraseLoRA, a dataset-free framework that replaces attention surgery with background-aware reasoning and test-time adaptation. The first stage, Background-aware Foreground Exclusion (BFE), leverages a multimodal large-language model to separate target foreground, non-target foregrounds, and clean background from a single image–mask pair. The second stage, Background-aware Reconstruction with Subtype Aggregation (BRSA), performs test-time optimization that treats inferred background subtypes as complementary pieces, enforcing their consistent integration through reconstruction and alignment objectives without explicit attention intervention. As a model-agnostic plug-in applicable to diverse diffusion backbones, EraseLoRA reconstructs backgrounds at least 23\% more faithful to the original scene than previous dataset-free methods while nearly halving unwanted foreground re-generation, and surpasses all dataset-driven approaches in both aspects despite requiring no training data. Code is available at \url{https://shjo-april.github.io/EraseLoRA}.
\end{abstract}

\begin{figure}[t]
  \centering
  \includegraphics[width=1.00\linewidth]{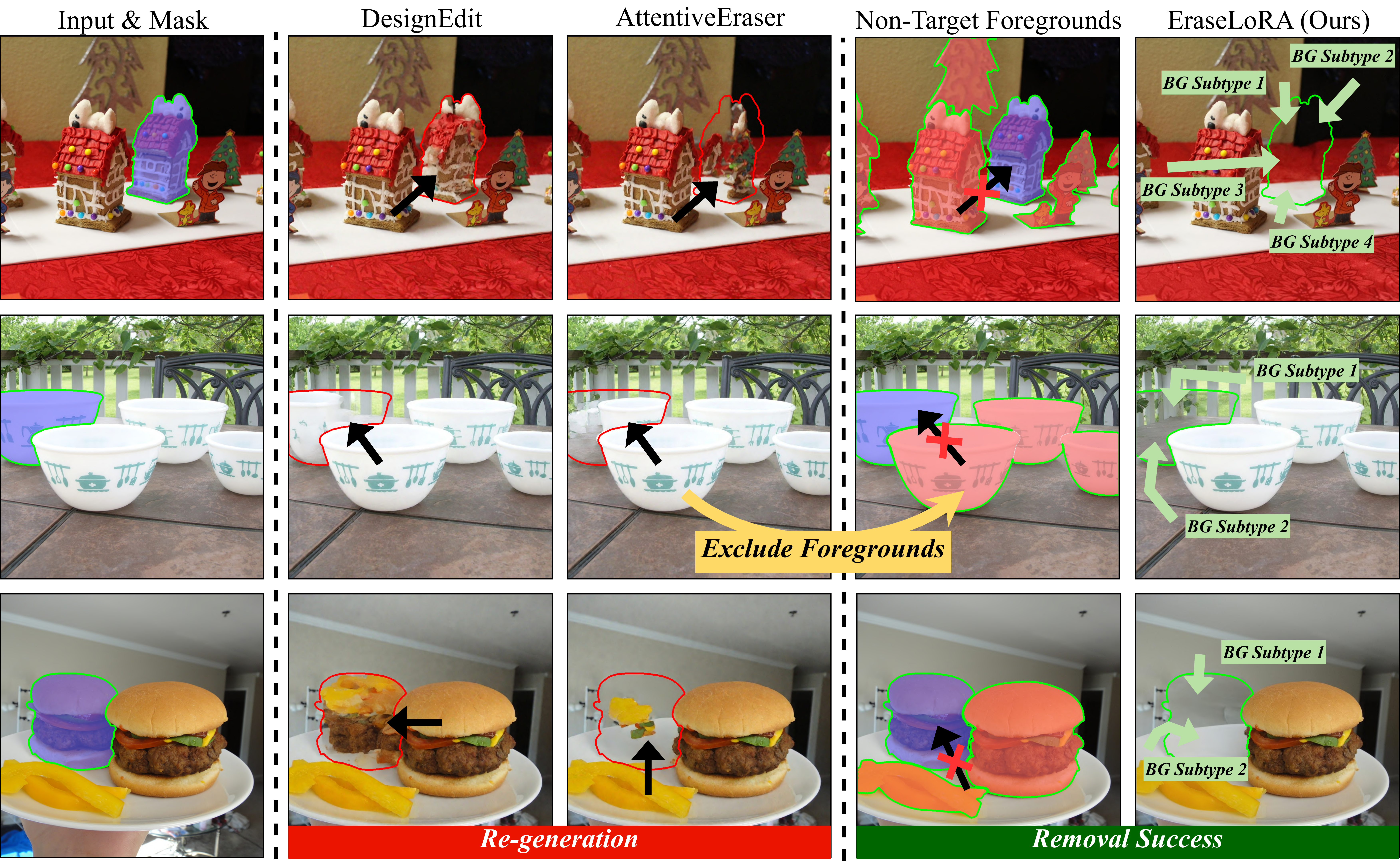}
  \captionof{figure}{
    \textbf{Qualitative comparison with prior dataset-free methods.} Previous state-of-the-art approaches \cite{jia2025designedit, sun2025attentiveeraser} treat only the masked region as foreground, misinterpreting non-target objects as background and regenerating them. EraseLoRA identifies and excludes non-target foregrounds and reconstructs the masked region using various background subtypes, enabling faithful object removal.
  }
  \label{fig:summary}
\end{figure}

\section{Introduction}
\label{sec:intro}

Image inpainting methods based on GANs \cite{Goodfellow2014gan, suvorov2022resolution, li2022mat} and text-to-image diffusion models \cite{ho2020ddpm, rombach2022sd, podell2024sdxl, esser2024dit} can synthesize visually plausible content in missing regions, yet they primarily aim to generate realistic textures rather than restore the underlying background structure, often hallucinating new objects instead of faithfully reconstructing what lies behind the removed target.

Object removal, in contrast, requires both eliminating the target and recovering the occluded background by transferring clean background cues into the masked region with structural consistency.
Recent dataset-free diffusion methods \cite{jia2025designedit, sun2025attentiveeraser} attempt to achieve this by redirecting or blocking self-attention within the masked region so that the model focuses on unmasked context. While effective in simple cases, these approaches share two inherent limitations. First, they treat the masked region as the only foreground and often misinterpret non-target foregrounds outside the mask as background, causing unintended regeneration of objects (see Fig.~\ref{fig:summary}). This reflects a lack of background-aware reasoning. 
Second, constraining attention uniformly without distinguishing distinct background subtypes compromises local detail and fails to coherently integrate multiple background cues, leading to blurred textures and unnatural boundaries between disparate subtypes (see Fig.~\ref{fig:artifact_attention}; further analysis in Appendix~\ref{subsec:attention_artifacts}).

\begin{figure}[t]
    \centering
    \includegraphics[width=\linewidth]{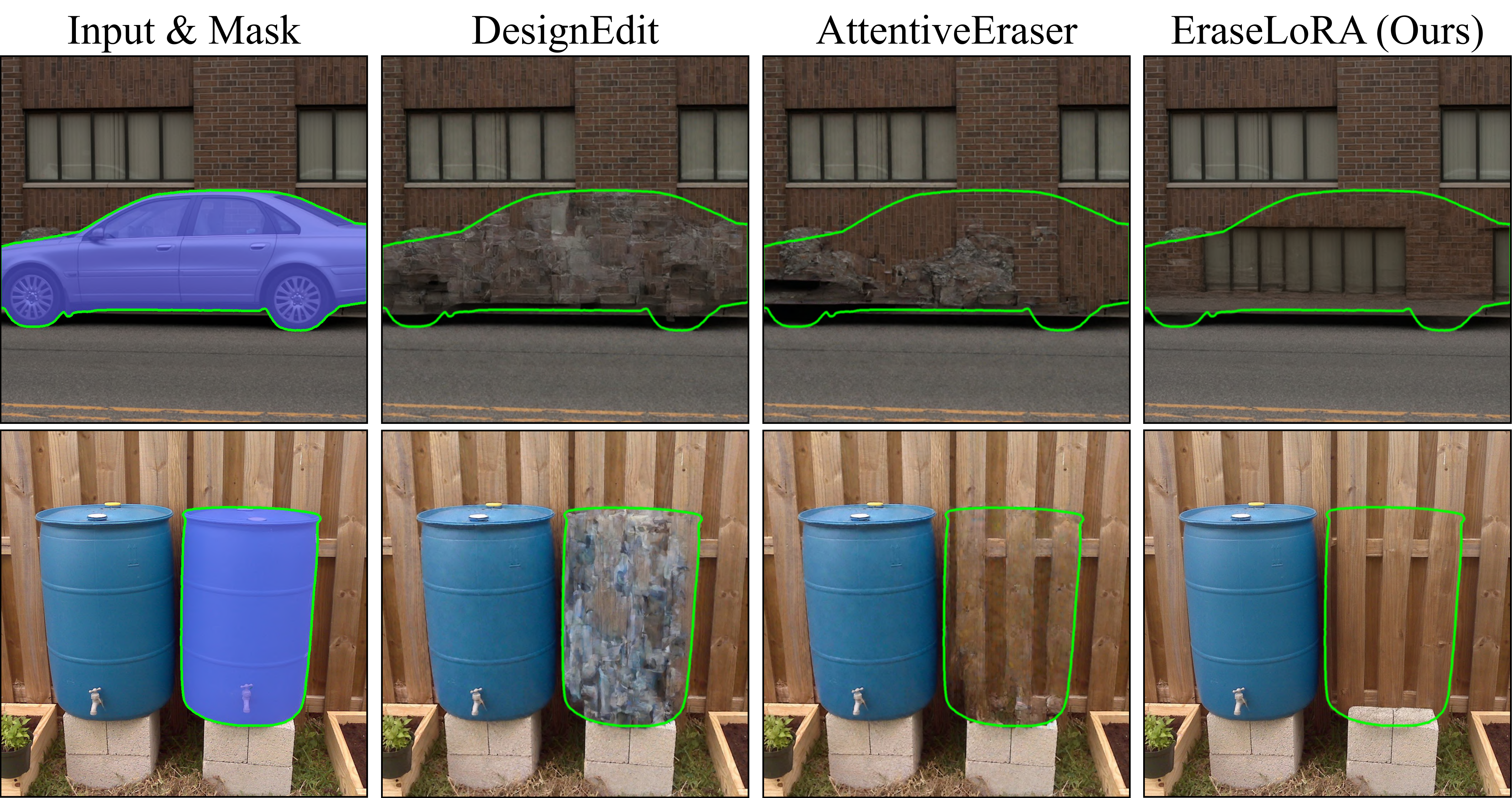} 
    \caption{\textbf{Artifacts from attention manipulation.} Recent dataset-free methods \cite{jia2025designedit, sun2025attentiveeraser} directly modify attention inside the mask without identifying background cues, leading to blurred or distorted textures, whereas EraseLoRA aggregates background subtypes without attention blocking and preserves sharp, coherent structures.}
    \label{fig:artifact_attention}
\end{figure}

We introduce EraseLoRA, a dataset-free framework that addresses these issues through background-aware reasoning and test-time adaptation. The first stage, Background-aware Foreground Exclusion (BFE), leverages a multimodal large-language model (MLLM) \cite{bai2025qwen2.5vl, zhu2025internvl3} to produce clean background cues by separating target foreground, non-target foregrounds, and background from a single image–mask pair. 
The second stage, Background-aware Reconstruction with Subtype Aggregation (BRSA), performs test-time optimization with Low-Rank Adaptation (LoRA) \cite{hu2022lora} to inject these cues into the masked area and aggregate multiple inferred background subtypes into a coherent reconstruction without a dataset-level removal prior or explicit attention blocking.
EraseLoRA is validated as a plug-in across diverse pretrained diffusion backbones \cite{esser2024dit, flux2024} and standard object-removal benchmarks \cite{kuznetsova2020openimagesv7, sagong2022rord}, demonstrating consistent gains over dataset-free baselines and competitive performance with dataset-driven methods. 
By combining the reasoning capability of MLLMs with the generative fidelity of diffusion models, EraseLoRA establishes an extensible formulation of dataset-free object removal that requires no additional data or retraining.

Our key contributions are as follows:
\begin{itemize}
    \item We identify a fundamental failure mode in object removal: non-target foregrounds are frequently misinterpreted as background, causing their unintended regeneration across recent dataset-free methods.
    \item We propose EraseLoRA, a background-aware, dataset-free object-removal framework that combines MLLM-guided separation of target and non-target foregrounds from background with a multi-background-aware test-time adaptation scheme, preventing foreground regeneration while maintaining contextual coherence.
    \item We provide three-label ground-truth annotations that distinguish target foreground, non-target foreground, and background, along with two evaluation metrics designed for unpaired object-removal settings.
    \item EraseLoRA improves background fidelity by at least 23\% over previous dataset-free methods while nearly halving unwanted foreground re-generation, and retains these gains when diffusion backbones and MLLMs are each replaced by alternatives.
\end{itemize}

\section{Related Work}
\label{sec:related_work}

\begin{table}[t]
    \centering
    \caption{Conceptual comparison of EraseLoRA (Ours) with previous approaches for object removal.}
    \label{tab:priorworks_comparison}
    \begin{adjustbox}{max width=\textwidth}
    {\large 
    \begin{tabular}{l|ccccccc} 
        \toprule
        \rowcolor{gray!15}
        {Properties} &
        \makecell{[ECCV'24] \\ PowerPaint \\ \cite{zhuang2024powerpaint}} &
        \makecell{[CVPR'25] \\ EntityErasure \\ \cite{zhu2025entityerasure}} &
        \makecell{[CVPR'25] \\ SmartEraser \\ \cite{jiang2025smarteraser}} &
        \makecell{[CVPR'26] \\ ObjectClear \\ \cite{zhao2026objectclear}} &
        \makecell{[AAAI'25] \\ DesignEdit \\ \cite{jia2025designedit}} &
        \makecell{[AAAI'25] \\ AttentiveEraser \\ \cite{sun2025attentiveeraser}} &
        \textbf{EraseLoRA} \\ 
        \midrule
        {Dataset-free object removal} &  
        \xmark & \xmark & \xmark & \xmark & \cmark & \cmark & \cmark \\
        \hline
        \makecell[l]{Identifies non-target \\ foregrounds with backgrounds} & 
        \xmark & \xmark & \xmark & \xmark & \xmark & \xmark & \textbf{\cmark} \\
        \hline
        \makecell[l]{Leverages multiple background subtypes} & 
        \xmark & \xmark & \xmark & \xmark & \xmark & \xmark & \textbf{\cmark} \\
        \hline
        \makecell[l]{Model-agnostic applicability} & 
        \xmark & \xmark & \xmark & \xmark & \cmark & \cmark & \textbf{\cmark} \\
        \bottomrule
    \end{tabular}
    }
    \end{adjustbox}
\end{table}

\subsection{Image Inpainting with Generative Models}
Image inpainting aims to complete missing regions using the visible context. Early approaches \cite{zhao2021comodgan, zuo2023ganseg, sargsyan2023migan} based on GANs have been surpassed by diffusion-based methods \cite{Xie2023smartbrush, yang2023paintbyexample}, which produce stable, detailed, and high-fidelity completions. Building on text-to-image diffusion backbones \cite{rombach2022sd, podell2024sdxl}, existing methods \cite{manukyan2025hdpainter, Xie2023smartbrush} fine-tune these backbones on paired inpainting datasets so that they can exploit text prompts while learning to fill masked regions. However, these approaches are still optimized for context-consistent completion and tend to hallucinate plausible new objects rather than preserving the original scene content.

\subsection{Diffusion Models for Object Removal}
Object removal is a specialized form of inpainting that must not only erase the masked target but also restore the occluded background with structural and contextual fidelity while preserving target-unrelated regions. Existing methods fall into two categories: dataset-driven approaches \cite{ekin2024clipaway, zhuang2024powerpaint, liu2025erasediff, jiang2025smarteraser}, which learn removal priors from additional, often paired before/after data, and dataset-free approaches \cite{chen2024freecompose, jia2025designedit, sun2025attentiveeraser}, which operate directly on pretrained text-to-image diffusion models \cite{rombach2022sd, podell2024sdxl, esser2024dit} without additional data. However, dataset-driven methods inherit a static training distribution and do not explicitly distinguish non-target foregrounds from background, which can leave object traces or perturb target-unrelated regions in complex scenes. Moreover, constructing paired examples is expensive and often unrealistic, since most pairs must be synthesized or extracted from video frames. These limitations motivate dataset-free methods that control the diffusion process at inference time.
Recent extensions address object effects such as shadows and reflections~\cite{wei2025omnieraser, fu2026EffectErase, zhao2026objectclear}. While related, they pursue a broader removal target and can distort target-unrelated regions, which object removal should preserve. We focus on removal within the given mask, prioritizing faithful restoration and preservation of target-unrelated regions.

Recent state-of-the-art dataset-free approaches \cite{jia2025designedit, sun2025attentiveeraser} redirect or suppress self-attention within the masked region to guide the model toward unmasked context. While this attention manipulation reduces unwanted regeneration to some extent, it introduces two fundamental limitations. First, these methods remain background-agnostic: by treating only the masked area as foreground, they often misinterpret non-target foregrounds as background and regenerate them, as shown in \cref{fig:summary}. Second, blocking or altering attention disrupts fine textures and prevents consistent integration of multiple background cues, leading to blurry or structurally inconsistent results, as illustrated in \cref{fig:artifact_attention}.

To address these limitations, we introduce a background-aware, dataset-free framework that leverages visual reasoning from an external model to identify and exclude non-target foregrounds and to produce clean background cues for reconstruction. We further aggregate multiple inferred background subtypes coherently through test-time adaptation, enabling structurally consistent background restoration without a dataset-level removal prior or explicit attention blocking. Our method is plug-and-play and model-agnostic, applicable across diverse diffusion backbones. Tab.~\ref{tab:priorworks_comparison} summarizes this design, highlighting how EraseLoRA differs from previous state-of-the-art methods \cite{zhu2025entityerasure, jiang2025smarteraser, jia2025designedit, sun2025attentiveeraser}.

\begin{figure}[t]
    \centering
    \includegraphics[width=1.00\linewidth]{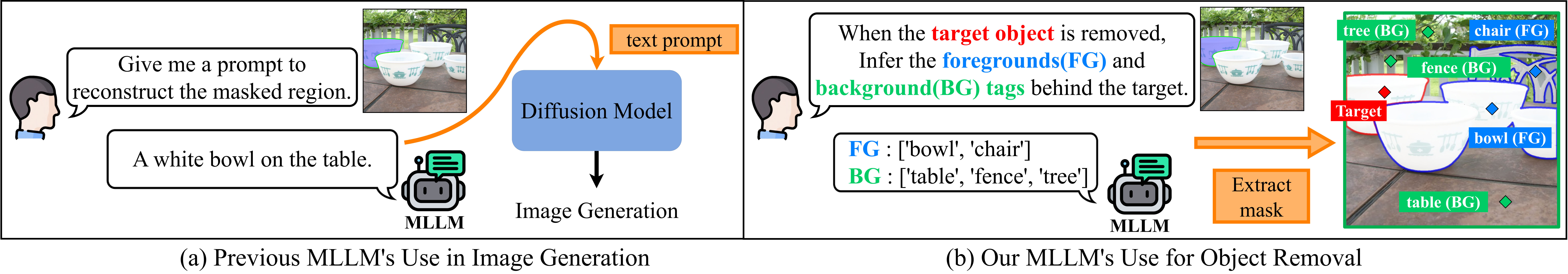} 
    \caption{\textbf{Background-aware reasoning power of MLLM.} Unlike prior works \cite{kim2025chainofzoom, wang2024genartist, Qu2025silmm, zhou2025fireedit} employ MLLMs for visual reasoning over the visible scene, we first leverage MLLMs to infer background cues behind the masked target.}
    \label{fig:mllm_use}
\end{figure}

\begin{figure}[t]
    \centering
    \includegraphics[width=1.00\linewidth]{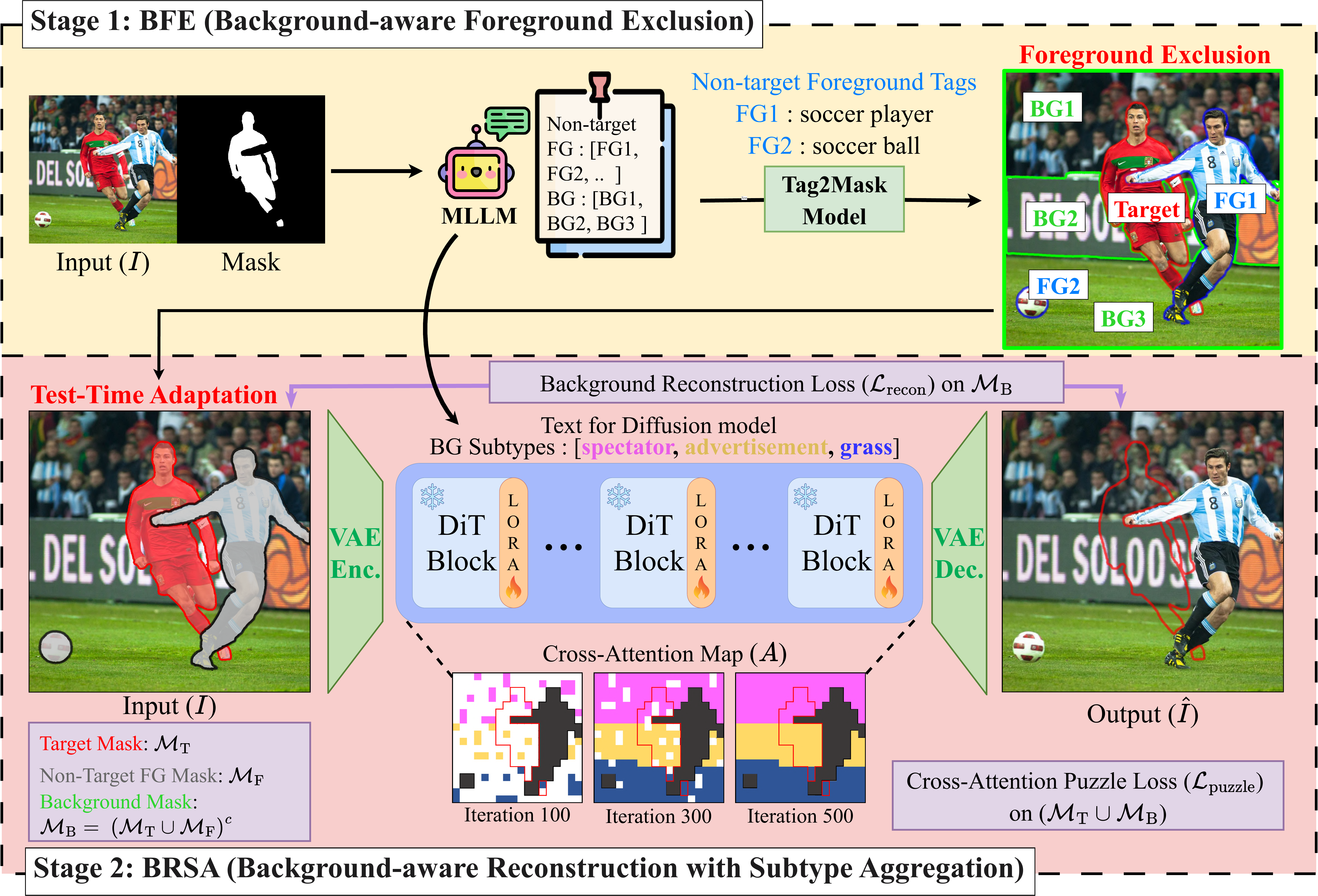} 
    \caption{{\textbf{Overview of EraseLoRA.} BFE (\cref{sec:bfe}) separates target foreground, non-target foregrounds, and background from a single image-mask pair using an MLLM \cite{zhu2025internvl3} and Tag2Mask models \cite{liu2024groundingdino, ravi2025sam2}. After producing clean background cues, BRSA (\cref{sec:brsa}) performs test-time adaptation \cite{wang2021tta} with reconstruction and alignment objectives, coherently integrating background subtypes into the masked region.}}
    \label{fig:overview}
\end{figure}

\subsection{MLLMs for Visual Reasoning in Image Editing}
Multimodal large-language models (MLLMs) \cite{liu2023llava, bai2025qwen2.5vl, bai2025qwen3, zhu2025internvl3} have gained traction in vision-language tasks due to their ability to interpret visual scenes and reason about object relations. Recent editing methods \cite{kim2025chainofzoom, wang2024genartist} leverage this capability to extract semantic descriptions, generate editing instructions, or guide global scene manipulation, while inpainting works \cite{fanelli2025idreammypainting, tianyidan2025anywhere, zhou2025fireedit} use MLLMs to analyze the visible context and propose content to fill masked regions. However, these approaches rely on visible context and aim to generate new objects, rather than inferring what lies behind a removed target.

We leverage MLLMs in a fundamentally different role: as background-aware reasoners for object removal. Rather than generating new foreground content, we use MLLMs to infer the occluded background behind the target and to identify non-target foregrounds that cause unintended regeneration, producing clean background cues that are not directly visible in the input image (see Fig.~\ref{fig:mllm_use}).

\section{Method}
\label{sec:method}

Our proposed EraseLoRA is a dataset-free object removal framework that leverages MLLM-guided background reasoning and test-time adaptation to achieve coherent background reconstruction. It consists of two stages: Background-aware Foreground Exclusion (BFE; Sec.~\ref{sec:bfe}) and Background-aware Reconstruction with Subtype Aggregation (BRSA; Sec.~\ref{sec:brsa}). The overall pipeline is illustrated in Fig.~\ref{fig:overview}, and we provide diffusion and attention preliminaries in Appendix~\ref{sec:preliminaries}.

\begin{figure}[t]
    \centering
    \includegraphics[width=1.00\linewidth]{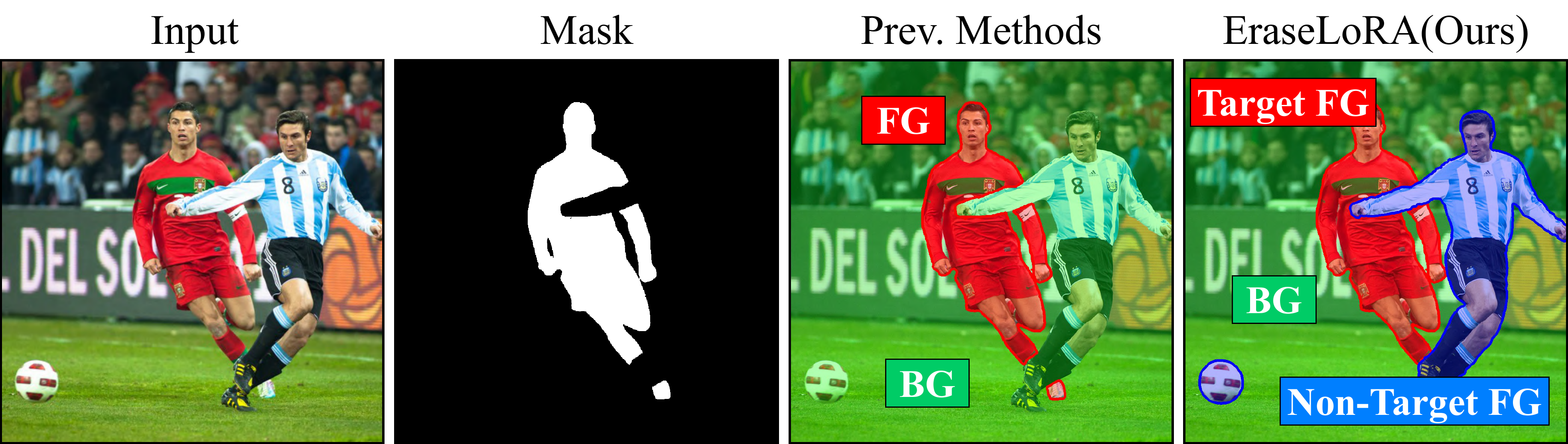} 
    \caption{\textbf{Identification of non-target foregrounds.} Prior methods \cite{chen2024freecompose, sun2025attentiveeraser} treat the entire unmasked region as background, which causes regeneration of non-target foregrounds. In contrast, EraseLoRA explicitly identifies non-target foregrounds within the unmasked region and excludes them, producing clean background.}
    \label{fig:3label_cls}
\end{figure}

\subsection{Background-aware Foreground Exclusion}
\label{sec:bfe}
The first stage, BFE, prevents unintended object regeneration by explicitly excluding non-target foregrounds from reference regions and extracting clean background cues for contextually coherent reconstruction.
Given an input image $I$ and a target mask $M_T$, we leverage the background-aware reasoning of MLLMs \cite{bai2025qwen2.5vl, zhu2025internvl3} to partition target foreground, non-target foregrounds, and background.
The MLLM first identifies all semantic tags in the image and classifies the masked object as the target foreground, visible objects that may cause regeneration as non-target foreground tags $\mathcal{F}$, and occluded objects or scene components behind the target as background subtype tags $\mathcal{B}$.
For each non-target foreground tag in $\mathcal{F}$, we use Tag2Mask models (\emph{e.g.}, Grounding DINO~\cite{liu2024groundingdino} and SAM2~\cite{ravi2025sam2}) to localize its corresponding region. The union of these localized regions defines the non-target foreground mask; the remaining pixels outside both the target and non-target foreground regions are treated as clean background.

We define three binary masks $M_T$, $M_F$, and $M_B$ over the latent spatial domain $\Omega=\{1,\dots,h\times w\}$. Following the VAE architecture of the employed diffusion backbone (\emph{e.g.}, \cite{esser2024dit}), we set $h{=}H/8$ and $w{=}W/8$ for an input of size $H \times W$:
\begin{equation}
\Omega = M_T \cup M_F \cup M_B ,
\label{eq:region_label}
\end{equation}
where $M_T$, $M_F$, and $M_B$ denote the target, non-target foreground, and clean background masks in the latent space, respectively.
This partition explicitly distinguishes non-target foreground objects from the unmasked background region, allowing us to isolate distractors that previous dataset-free methods \cite{sun2025attentiveeraser, jia2025designedit} mistakenly treat as background (see Fig.~\ref{fig:3label_cls}), which in turn yields cleaner background supervision for subsequent reconstruction.

\subsection{Background-aware Reconstruction with Subtype Aggregation}
\label{sec:brsa}
Based on the clean background cues obtained in BFE, BRSA performs test-time optimization with Low-Rank Adaptation (LoRA) \cite{hu2022lora} to effectively aggregate multiple background subtypes and reconstruct the masked region with structural and contextual consistency. 
This adaptation provides image-specific background fidelity in dataset-free setting without artifacts from attention manipulation.
To achieve this, BRSA jointly optimizes two complementary objectives: the Background Reconstruction Loss ($\mathcal{L}_{\text{recon}}$) and the Background Puzzle Loss ($\mathcal{L}_{\text{puzzle}}$). 
The overall objective is formulated as
$\mathcal{L}_{\text{total}} = \mathcal{L}_{\text{recon}} + \lambda \, \mathcal{L}_{\text{puzzle}}$, where $\lambda{=}0.2$ balances background reconstruction fidelity and subtype aggregation.
Together, these losses guide how background information is integrated into the masked region, softly regulating attention flow without the hard attention-blocking used in prior methods~\cite{jia2025designedit, sun2025attentiveeraser}.

\begin{figure}[t]
    \centering
    \includegraphics[width=1.00\linewidth]{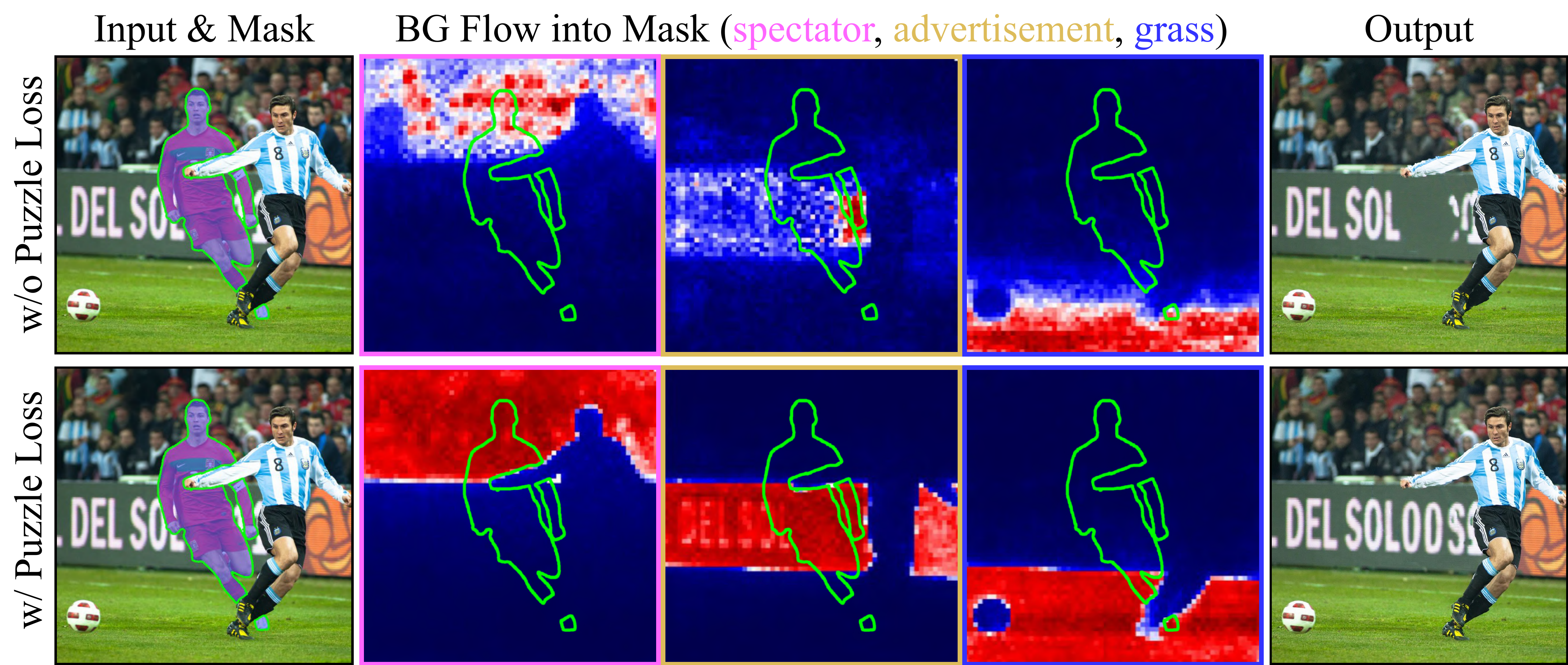} 
    \caption{\textbf{Effect of the background puzzle loss.} We visualize how each background subtype (spectator, advertisement, grass) is represented inside the mask. The background puzzle loss ensures structurally coherent integration of background subtypes within the mask, unlike the weak integration without it.}
    \label{fig:puzzle_loss}
\end{figure}

\textbf{Background Reconstruction Loss.} To preserve regions that are confidently identified as clean background by BFE (\cref{sec:bfe}), we impose a reconstruction loss only on $M_B$. Let $z = \mathrm{Enc}(I)$ be the latent representation of the input image $I$, and $\hat{z}$ be the reconstructed latent after denoising. The background reconstruction loss is defined as
\begin{equation}
\mathcal{L}_{\text{recon}}
=
\frac{1}{|M_B|}
\sum_{p \in M_B}
\big\| \hat{z}[p] - z[p] \big\|_2^2,
\label{eq:recon}
\end{equation}
where $p$ denotes spatial indices in the latent feature map. By anchoring $\hat{z}$ to $z$ on these background locations, $\mathcal{L}_{\text{recon}}$ enforces fidelity to the original background and promotes globally coherent reconstruction.

\textbf{Background Puzzle Loss.} While the reconstruction loss $\mathcal{L}_{\text{recon}}$ preserves high-fidelity background appearance, it does not explicitly control how different background subtypes are filled and integrated into the masked area, often leading to structurally inconsistent or partially missing patterns (\emph{e.g.}, misaligned background context in Fig.~\ref{fig:puzzle_loss}). To address this, we introduce a background puzzle loss that treats each background subtype as a distinct puzzle piece that must contribute coherently to reconstructing the masked region. This loss enforces that background attention is concentrated only on valid regions (target foreground or clean background), preventing attention distraction toward non-target foregrounds and improving spatial consistency:
\begin{equation}
\mathcal{L}_{\text{puzzle}}
=
1 - \mathrm{Dice}\!\left(
A^{\mathrm{dom}},
\mathbf{1}_{M_T \cup M_B}
\right),
\label{eq:puzzle}
\end{equation}
where $\mathrm{Dice}(\cdot,\cdot)$ computes the soft spatial overlap between the continuous attention map and the binary valid-region indicator (see Appendix~\ref{sec:details_of_method} for details). $A^{\mathrm{dom}}[p] = \max_{b \in \mathcal{B}} A_b[p]$ selects the strongest attention response at each location $p$ across background subtype tags $b \in \mathcal{B}$ inferred by BFE~(\cref{sec:bfe}), ensuring that every position is accounted for by its most relevant subtype.

\section{Experiments}
\label{sec:experiment}

\begin{figure}[t]
    \centering
    \includegraphics[width=1.00\linewidth]{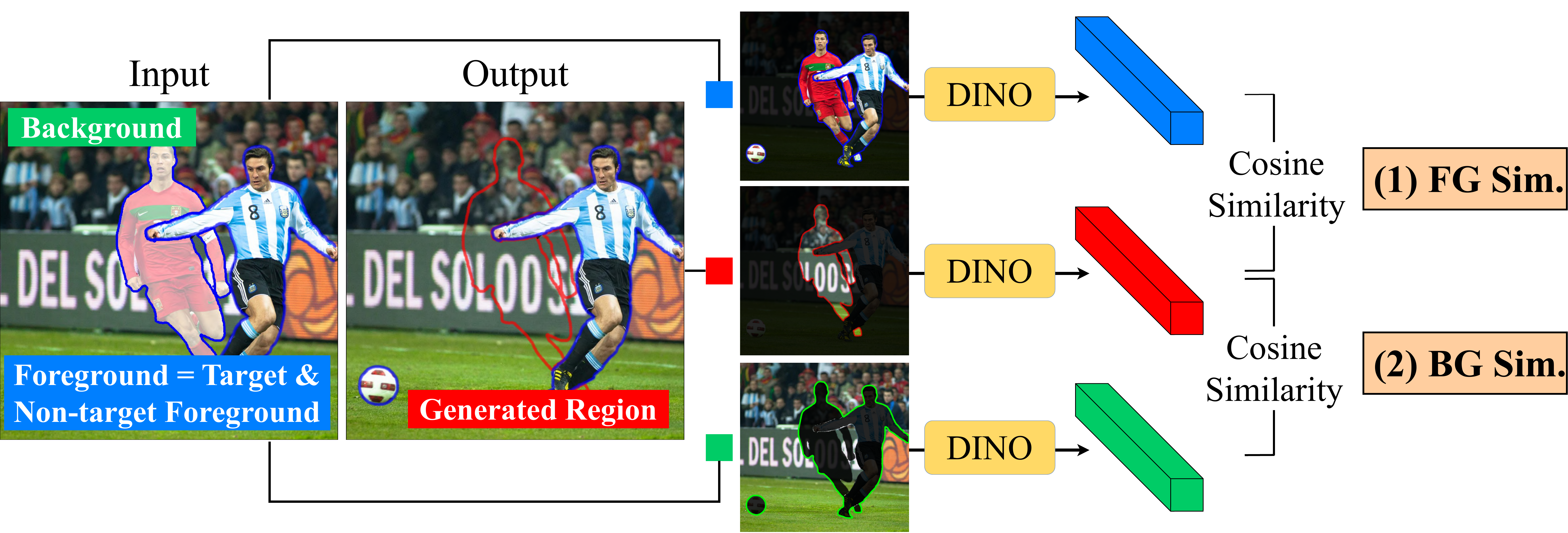} 
    \caption{Illustration of evaluation metrics (\emph{i.e.}, BG Sim. and FG Sim.) for unpaired object removal.}
    \label{fig:metrics}
\end{figure}

\subsection{Experimental Setup}

\textbf{Implementation details.} We implement EraseLoRA on three text-to-image diffusion backbones \cite{podell2024sdxl, esser2024dit, flux2024}. For a fair comparison, we strictly follow the official inference configurations (scheduler, guidance scale, resolution, and the number of sampling steps) and apply EraseLoRA in a purely dataset-free, test-time manner. During test-time adaptation (TTA), we freeze all backbone parameters and optimize only the inserted LoRA adapters \cite{hu2022lora} (rank $r{=}32$) for 500 iterations with respect to the final loss defined in \cref{sec:brsa}. This configuration is kept consistent across all backbones and benchmarks. For BFE (\cref{sec:bfe}), we use InternVL3-78B~\cite{zhu2025internvl3} as the default MLLM and Grounded SAM2~\cite{ren2024groundedsam, ravi2025sam2} for tag-to-mask conversion. Additional details are provided in Appendix~\ref{sec:details_of_method}.

\textbf{Benchmarks.} We evaluate EraseLoRA on two benchmarks: 200 samples from OpenImages~V7~\cite{kuznetsova2020openimagesv7} and 343 frames from RORD~\cite{sagong2022rord}. In both datasets, the original annotations do not distinguish non-target foregrounds from background, so we annotate them with three-label (target / non-target foreground / background) ground-truth masks to capture distractor regions. These refined annotations will be released and form the basis of the evaluation metrics described below. Additional results on RemovalBench~\cite{wei2025omnieraser} are provided in Appendix~\ref{sec:appendix_quantitative_results}.
 
\textbf{Evaluation metrics.} 
For unpaired object removal, we use DINO similarity~\cite{caron2021dino}, following recent image generation and editing works~\cite{li2024tuning, zhang2023magicbrush}. Foreground Similarity (FG Sim.) and Background Similarity (BG Sim.) measure, respectively, how much the reconstructed region stays similar to the foreground (lower is better) and how well it aligns with the background (higher is better; see Fig.~\ref{fig:metrics}). 
We additionally report Background Preservation (BG Pres.), computed via SSIM~\cite{wang2004ssim} on unmasked regions, following the evaluation protocol of recent editing methods~\cite{zhu2025kv, kim2025early}, to assess fidelity outside the masked area. 

\begin{table}[t]
   \centering
   \caption{Quantitative comparison with previous state-of-the-art methods on test datasets \cite{kuznetsova2020openimagesv7, sagong2022rord}. The best results are in \textbf{bold}.}
   \label{tab:quantitative_results}
   \resizebox{\linewidth}{!}{
   \begin{tabular}{lcccccc}
   \toprule
   \multirow{3}{*}{ } & \multicolumn{3}{c}{\textbf{OpenImages V7}} & \multicolumn{3}{c}{\textbf{RORD}} \\
   \cmidrule(lr){2-4} \cmidrule(lr){5-7}
    & \textbf{BG Sim.($\uparrow$)} & \textbf{FG Sim.($\downarrow$)} & \textbf{BG Pres.($\uparrow$)} & \textbf{BG Sim.($\uparrow$)} & \textbf{FG Sim.($\downarrow$)} & \textbf{BG Pres.($\uparrow$)} \\
   \midrule
   \rowcolor{gray!15}
   \multicolumn{7}{l}{\textbf{\textit{Dataset-Free Approaches:}}} \\
   \midrule
   SD3.5-M~\cite{esser2024dit} & 0.605 & 0.286 & \textbf{0.934} & 0.582 & 0.319 & {0.907} \\
   + AttentiveEraser~\cite{sun2025attentiveeraser} & 0.559 & 0.276 & 0.931 & 0.541 & 0.302 & 0.901 \\
   + DesignEdit~\cite{jia2025designedit} & 0.600 & 0.255 & {0.933} & 0.597 & 0.273 & \textbf{0.908} \\
   \rowcolor{green!15}
    \textbf{+ EraseLoRA} & \textbf{0.743} & \textbf{0.151} & 0.924 & \textbf{0.779} & \textbf{0.138} & {0.901} \\
   \midrule
   \rowcolor{gray!15}
   \multicolumn{7}{l}{\textbf{\textit{Dataset-Driven Approaches:}}} \\
   \midrule
   SDXL-Inpainting~\cite{podell2024sdxl} & 0.677 & 0.212 & 0.742 & 0.645 & 0.234 & 0.720 \\
   PowerPaint~\cite{zhuang2024powerpaint} & 0.669 & 0.217 & 0.719 & 0.729 & 0.176 & 0.687 \\
   CLIPAway~\cite{ekin2024clipaway} & 0.656 & 0.223 & 0.713 & 0.744 & 0.156 & 0.705 \\
   SmartEraser~\cite{jiang2025smarteraser} & {0.709} & {0.185} & 0.727 & {0.768} & {0.148} & 0.672 \\
   EntityErasure~\cite{zhu2025entityerasure} & 0.679 & 0.204 & 0.728 & 0.766 & 0.175 & 0.716 \\
   \bottomrule
   \end{tabular}
    }   
\end{table}

\subsection{Comparison with State-of-the-art Approaches}
\begin{figure}[t]
    \centering
    \includegraphics[width=\linewidth]{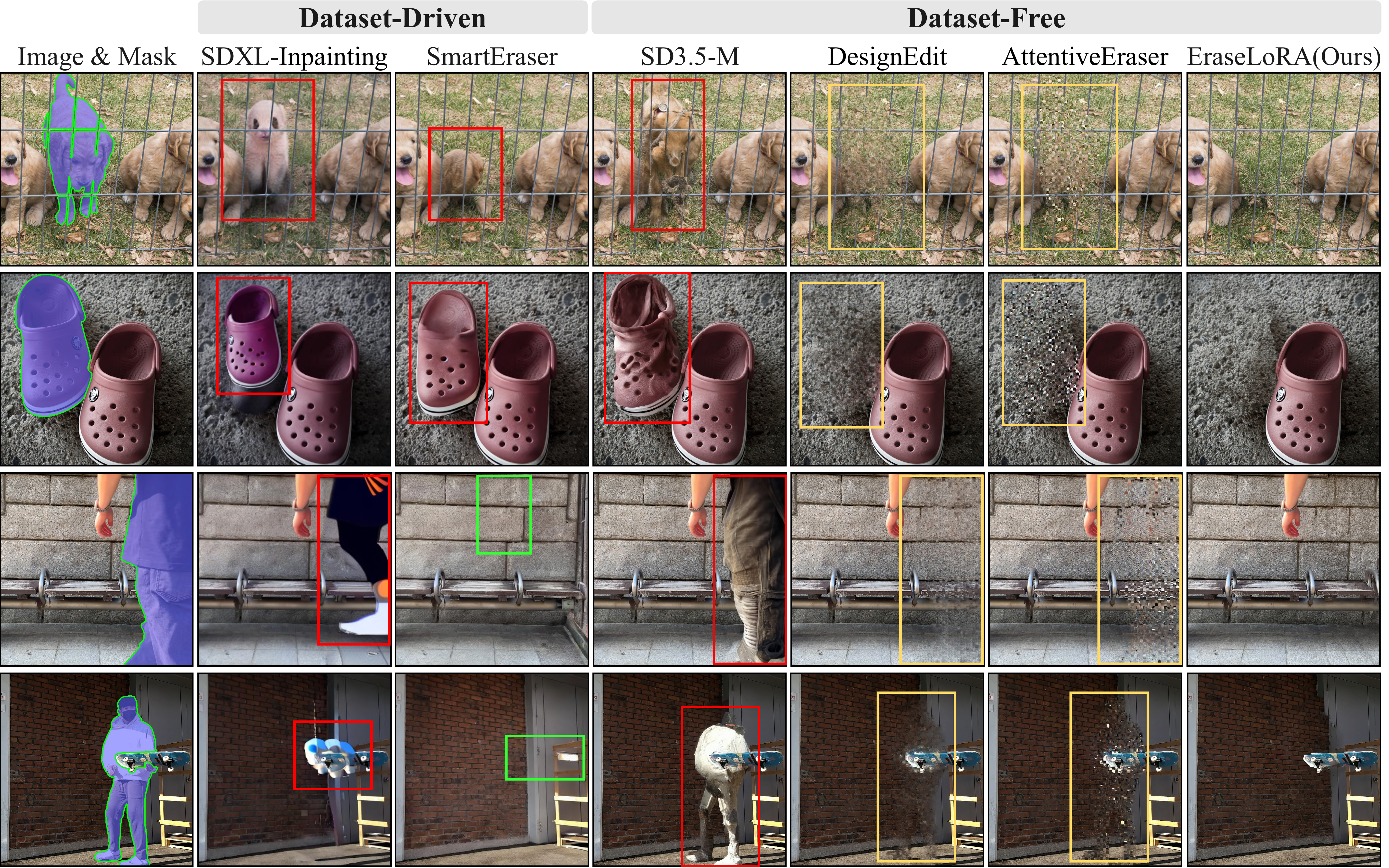} 
    \caption{Qualitative comparison on OpenImages~V7~\cite{kuznetsova2020openimagesv7} and RORD~\cite{sagong2022rord}. Without any paired data, EraseLoRA avoids unwanted background changes (green), copying nearby objects (red), and residual foreground artifacts (yellow), achieving cleaner object removal and more faithful backgrounds than both dataset-driven and dataset-free methods~\cite{podell2024sdxl, jiang2025smarteraser, esser2024dit, jia2025designedit, sun2025attentiveeraser}.}
    \label{fig:qualitative_results}
\end{figure}

Table~\ref{tab:quantitative_results} shows the quantitative results on two benchmarks \cite{kuznetsova2020openimagesv7, sagong2022rord}. Compared to the baseline \cite{esser2024dit}, EraseLoRA substantially improves BG Sim., from 0.605 to 0.743 on OpenImages~V7~\cite{kuznetsova2020openimagesv7} and from 0.582 to 0.779 on RORD~\cite{sagong2022rord} (absolute gains of +0.14 and +0.20, corresponding to roughly 23\% and 34\% relative improvements). At the same time, FG Sim. is almost halved (0.286$\rightarrow$0.151 on OpenImages~V7, 0.319$\rightarrow$0.138 on RORD), indicating that EraseLoRA suppresses foreground re-generation while filling the mask with background-consistent content. 
EraseLoRA also outperforms dataset-driven methods~\cite{podell2024sdxl, zhuang2024powerpaint, ekin2024clipaway, jiang2025smarteraser, zhu2025entityerasure}: it attains the highest BG Sim. and the lowest FG Sim. on both benchmarks~\cite{kuznetsova2020openimagesv7, sagong2022rord}, while maintaining background preservation around 0.90, which is about 0.18 higher than all five dataset-driven methods. This indicates that EraseLoRA predominantly modifies only the masked region and leaves the unmasked background almost unchanged, unlike dataset-driven models that often perturb surrounding content. 
Qualitative comparisons in Fig.~\ref{fig:qualitative_results} reflect the same trend: EraseLoRA removes the target object cleanly while preserving sharp background details and avoiding unwanted edits in unmasked regions. Additional quantitative and qualitative results are provided in Appendix~\ref{sec:appendix_quantitative_results} and \ref{sec:appendix_qualitative_results}.

\subsection{Discussion}
To understand EraseLoRA’s performance gains, we conduct component-wise ablations on OpenImages~V7~\cite{kuznetsova2020openimagesv7}, measuring the effect of each stage and loss term.

\begin{figure}[t]
    \centering
    \includegraphics[width=\linewidth]{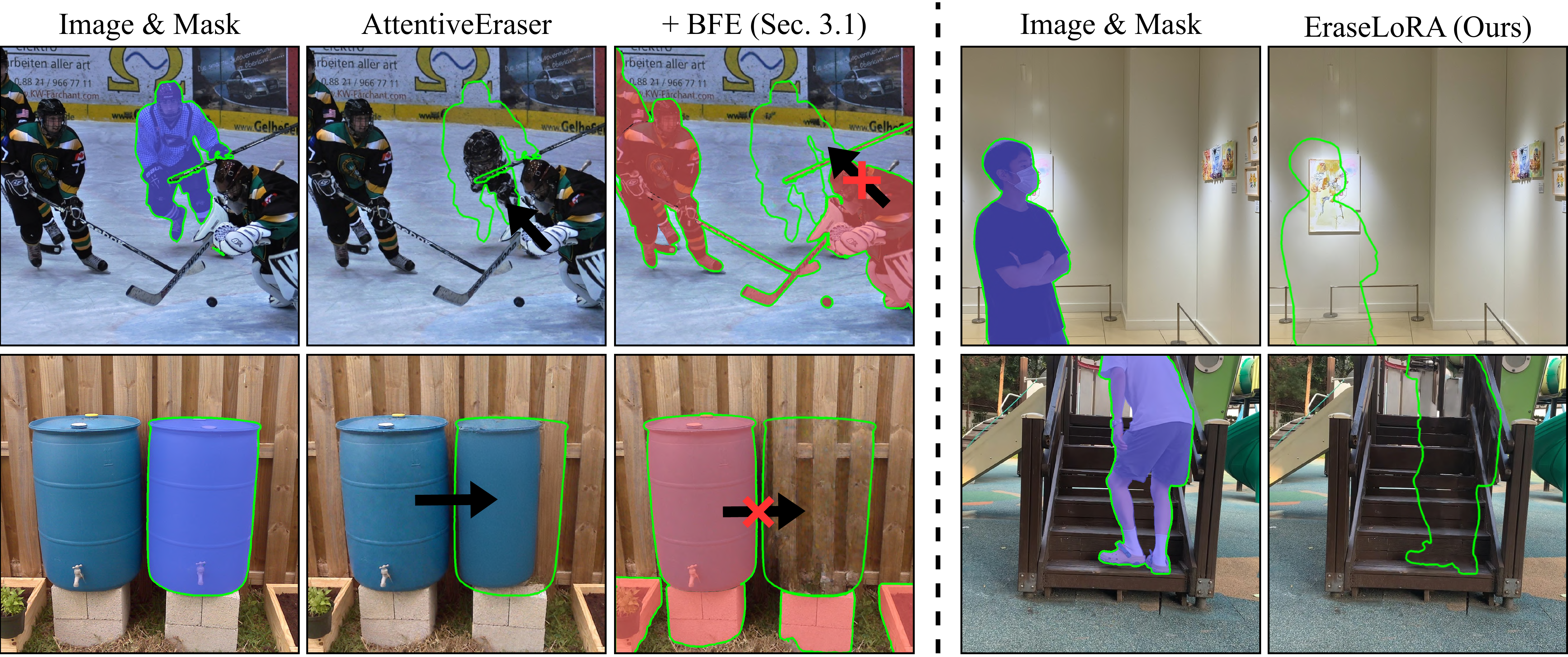} 
    \caption{(Left) Effect of BFE (\cref{sec:bfe}) on AttentiveEraser~\cite{sun2025attentiveeraser}. Red masks denote excluded non-target foregrounds. (Right) Examples of EraseLoRA in occlusion cases. EraseLoRA reconstructs occluded content (\emph{e.g.}, a painting, stairs) as background from surrounding context.}
    \label{fig:ab_bfe_occlusion}
\end{figure}

\textbf{Effect of BFE with prior dataset-free methods.} To test whether our foreground exclusion module (BFE; \cref{sec:bfe}) generalizes beyond EraseLoRA, we plug it into prior dataset-free object-removal methods~\cite{jia2025designedit, sun2025attentiveeraser} without modifying their inference pipelines. By explicitly excluding non-target foregrounds from their reference regions, background similarity improves by up to 6.6\%, while foreground similarity decreases by 8.6\% (\cref{tab:ab_bfe_brsa_effect}, left). As shown in the left panel of Fig.~\ref{fig:ab_bfe_occlusion}, non-target foregrounds that were previously regenerated are successfully suppressed once BFE provides clean background references, confirming that BFE alleviates the re-generation limitation of existing dataset-free approaches~\cite{jia2025designedit, sun2025attentiveeraser} in a fully model-agnostic manner.

\begin{table}[t]
    \centering
    \caption{(Left) Effect of non-target foreground exclusion (BFE; \cref{sec:bfe}) in previous state-of-the-art dataset-free methods \cite{jia2025designedit, sun2025attentiveeraser} and (Right) loss component in BRSA (\cref{sec:brsa}).}
    \label{tab:ab_bfe_brsa_effect}
    \resizebox{!}{1.17cm}{
        \begin{tabular}{p{3.5cm}cc}
            \toprule
            \multirow{2}{*}{\textbf{Method}} & \multicolumn{2}{c}{\textbf{Metrics}} \\ 
            \cmidrule(lr){2-3}
            & \textbf{BG Sim.($\uparrow$)} & \textbf{FG Sim.($\downarrow$)} \\
            \midrule
            AttentiveEraser~\cite{sun2025attentiveeraser} & 0.559 & 0.276 \\
            \rowcolor{green!15}
            + BFE (Ours; \cref{sec:bfe}) & \textbf{0.596} & \textbf{0.252} \\
            \midrule
            DesignEdit~\cite{jia2025designedit} & 0.600 & 0.255 \\
            \rowcolor{green!15}
            + BFE (Ours; \cref{sec:bfe}) & \textbf{0.603} & \textbf{0.251} \\
            \bottomrule
        \end{tabular}
    }
    \resizebox{!}{1.17cm}{ 
        \begin{tabular}{lcccc}
            \toprule
            \multirow{2}{*}{\textbf{Method}} & \multicolumn{2}{c}{\textbf{Loss Components}} & \multicolumn{2}{c}{\textbf{Metrics}} \\ 
            \cmidrule(lr){2-3} \cmidrule(lr){4-5}
            & $\mathcal{L}_{\text{recon}}$ & $\mathcal{L}_{\text{puzzle}}$ & \textbf{BG Sim.($\uparrow$)} & \textbf{FG Sim.($\downarrow$)} \\
            \midrule
            SD3.5-M~\cite{esser2024dit} & \xmark & \xmark & 0.605 & 0.286 \\
            + EraseLoRA & \cmark & \xmark & 0.736 & 0.158 \\
            + EraseLoRA & \xmark & \cmark & 0.561 & 0.278 \\
            \rowcolor{green!15}
            + EraseLoRA & \cmark & \cmark & \textbf{0.743} & \textbf{0.151} \\
            \bottomrule
        \end{tabular}
    }
\end{table}

\begin{figure}[t]
    \centering
    \includegraphics[width=\linewidth]{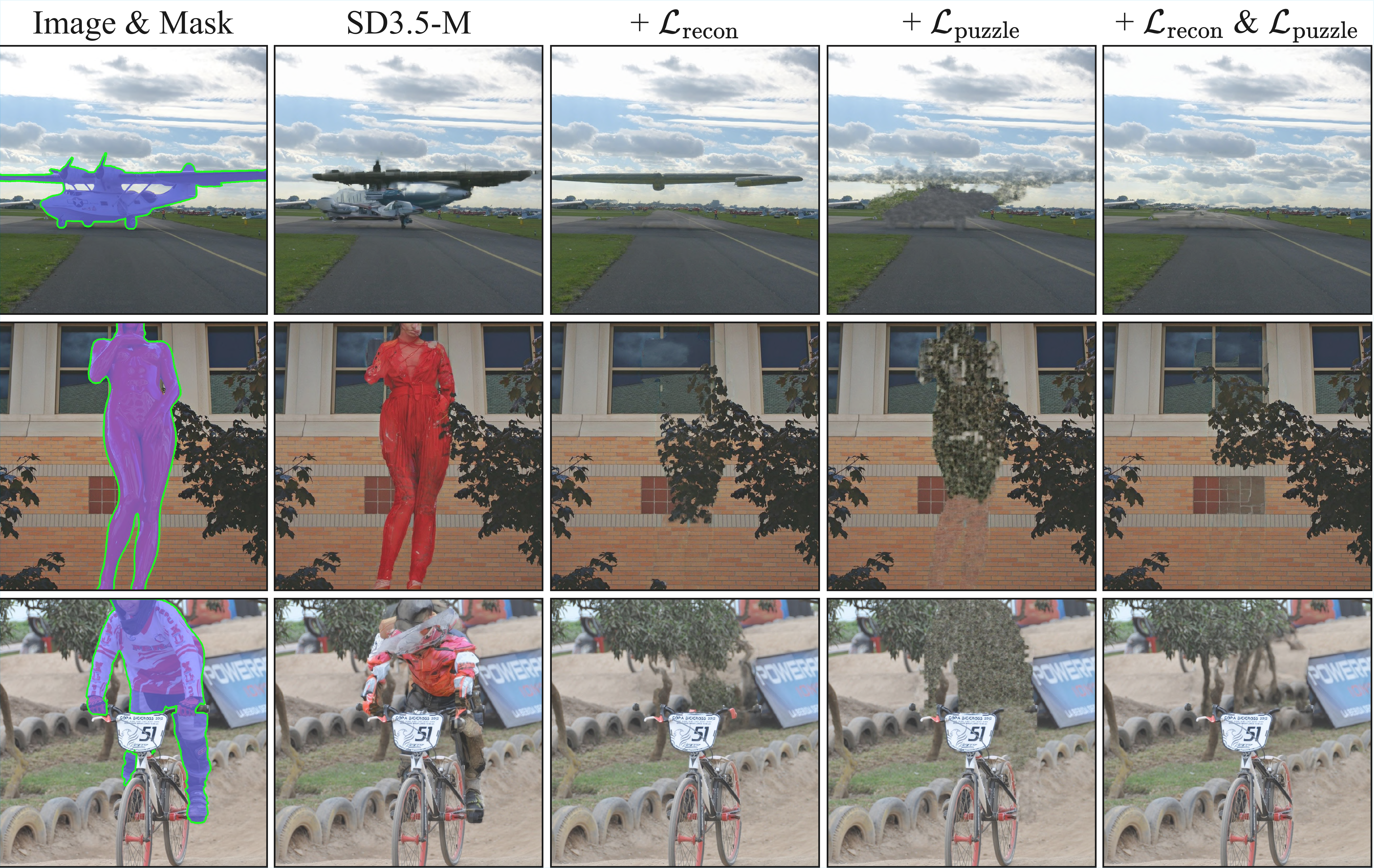} 
    \caption{Visualization of loss components in BRSA (\cref{sec:brsa}). $\mathcal{L}_{\text{recon}}$ preserves background structure and $\mathcal{L}_{\text{puzzle}}$ completes the masked region; using both (\emph{i.e.}, EraseLoRA) produces the most coherent and detailed background.}
    \label{fig:ab_loss_components}
\end{figure}

\textbf{Effect of losses in BRSA.} To determine the most effective objective for BRSA (\cref{sec:brsa}), we examine the roles of the two complementary losses through a loss-combination study (see Tab.~\ref{tab:ab_bfe_brsa_effect}, right). 
The background reconstruction loss ($\mathcal{L}_{\text{recon}}$; Eq.~\eqref{eq:recon}) preserves structural background consistency, improving BG Sim.\ from 0.605 to 0.736 (+21.7\%), but can still leave faint foreground traces.
In contrast, the background puzzle loss ($\mathcal{L}_{\text{puzzle}}$; Eq.~\eqref{eq:puzzle}) suppresses foreground artifacts via background subtype aggregation, but alone lacks explicit background anchoring and fails to capture fine background structure and global patterns, often leading to visually inconsistent completions (see Fig.~\ref{fig:ab_loss_components}).
Together (\emph{i.e.}, EraseLoRA), the two losses achieve the best results, yielding coherent and detail-preserving background restoration while overcoming the limitations observed when either loss is used alone.

\begin{table}[t]
    \centering
    \caption{Applicability across different (Left) MLLMs and (Right) Tag2Mask models.}
    \label{tab:ab_robustness_mllms_tag2mask}
    \resizebox{!}{1.14cm}{ 
        \begin{tabular}{lccc}
            \toprule
            \multirow{2}{*}{\textbf{Method}} & \multirow{2}{*}{\textbf{MLLM}} & \multicolumn{2}{c}{\textbf{Metrics}} \\ 
            \cmidrule(lr){3-4}
            & & \textbf{BG Sim.($\uparrow$)} & \textbf{FG Sim.($\downarrow$)} \\
            \midrule
            SD3.5-M~\cite{esser2024dit} & N/A & 0.605 & 0.286 \\
            + EraseLoRA & LLaVA-7B \cite{liu2023llava} & 0.728 & 0.164 \\
            + EraseLoRA & LLaVA-7B+MARINE \cite{zhao2025marine} & 0.727 & 0.161 \\
            + EraseLoRA & Qwen2.5-VL-72B \cite{bai2025qwen2.5vl} & 0.726 & 0.165 \\
            \rowcolor{green!15}
            + EraseLoRA & InternVL3-78B \cite{zhu2025internvl3} & \textbf{0.743} & \textbf{0.151} \\
            \bottomrule
        \end{tabular}
    }
    \resizebox{!}{1.14cm}{ 
        \begin{tabular}{lccc}
            \toprule
            \multirow{2}{*}{\textbf{Method}} & \multicolumn{1}{c}{\textbf{Tag2Mask}} & \multicolumn{2}{c}{\textbf{Metrics}} \\ 
            \cmidrule(lr){3-4}
            & \textbf{Model} & \textbf{BG Sim.($\uparrow$)} & \textbf{FG Sim.($\downarrow$)} \\
            \midrule
            SD3.5-M~\cite{esser2024dit} & N/A & 0.605 & 0.286 \\
            + EraseLoRA & Seg4Diff \cite{kim2025seg4diff} & 0.666 & 0.205 \\
            + EraseLoRA & YOLOE \cite{wang2025yoloe} & 0.709 & 0.177 \\
            + EraseLoRA & SAM3 \cite{carion2025sam3} & 0.708 & 0.177 \\
            \rowcolor{green!15}
            + EraseLoRA & G.SAM2 \cite{ren2024groundedsam} & \textbf{0.743} & \textbf{0.151} \\
            \bottomrule
        \end{tabular}
    }
\end{table}

\textbf{Flexibility.}
To evaluate the general applicability of EraseLoRA, we apply our method to different text-to-image diffusion backbones~\cite{rombach2022sd, podell2024sdxl, esser2024dit, flux2024}, MLLMs of varying scales \cite{liu2023llava, zhao2025marine, bai2025qwen2.5vl, zhu2025internvl3}, and Tag2Mask models \cite{kim2025seg4diff, wang2025yoloe, carion2025sam3, ren2024groundedsam}.
EraseLoRA demonstrates generalization ability by yielding reliable gains in both BG Sim. and FG Sim. regardless of the underlying diffusion backbones (see Tab.~\ref{tab:appendix_robustness_backbones} and Fig.~\ref{fig:appendix_robustness_backbones}). Detailed quantitative and qualitative analyses are provided in Appendix~\ref{subsec:appendix_flexibility}.

We also evaluate EraseLoRA across MLLMs of various scales~\cite{liu2023llava, zhao2025marine, bai2025qwen2.5vl, zhu2025internvl3} from 7B to 78B parameters, including a hallucination-mitigated variant~\cite{zhao2025marine}. Across all these models, EraseLoRA provides consistent gains (Tab.~\ref{tab:ab_robustness_mllms_tag2mask}, left), confirming that our framework reliably leverages the background-aware reasoning ability of MLLMs rather than depending on a specific architecture. 
Notably, even lightweight 7B MLLMs yield substantial improvements comparable to much larger models~\cite{bai2025qwen2.5vl}, achieving gains of up to 20.3\% in BG Sim. and 42.7\% in FG Sim., which indicates that EraseLoRA remains highly effective in resource-constrained settings without requiring a large MLLM.
Despite their effectiveness, we adopt a large-scale MLLM~\cite{zhu2025internvl3} as our default configuration in the main experiments, as it provides the strongest background-aware reasoning and achieves the best overall performance. 

We further validate EraseLoRA’s effectiveness across different Tag2Mask models~\cite{kim2025seg4diff, wang2025yoloe, carion2025sam3, ren2024groundedsam} in BFE (\cref{sec:bfe}). 
Across all Tag2Mask variants, EraseLoRA consistently improves background reconstruction and foreground suppression, yielding at least 10.0\% gains in BG Sim. and 28.3\% reductions in FG Sim. over the SD3.5-M~\cite{esser2024dit} baseline (see Tab.~\ref{tab:ab_robustness_mllms_tag2mask}, right).
Grounded SAM2 achieves the best performance, improving BG Sim. by up to 22.8\% and reducing FG Sim. by up to 47.2\%, resulting in the cleanest and most faithful background reconstruction. Additional qualitative results and detailed analysis regarding the impact of MLLM hallucinations are provided in Appendix~\ref{subsec:appendix_flexibility} and \ref{subsec:appendix_discussion}.

\textbf{MLLM-guided removal under occlusion.}
Occlusion poses a unique challenge for object removal: content normally regarded as foreground may be hidden behind the target, yet must be recovered as background during inpainting. By leveraging the MLLM's scene-level understanding, EraseLoRA correctly identifies such occluded elements and reconstructs them with semantically coherent completions (see Fig.~\ref{fig:ab_bfe_occlusion}, right).

\begin{figure}[t]
    \centering
    \includegraphics[width=\linewidth]{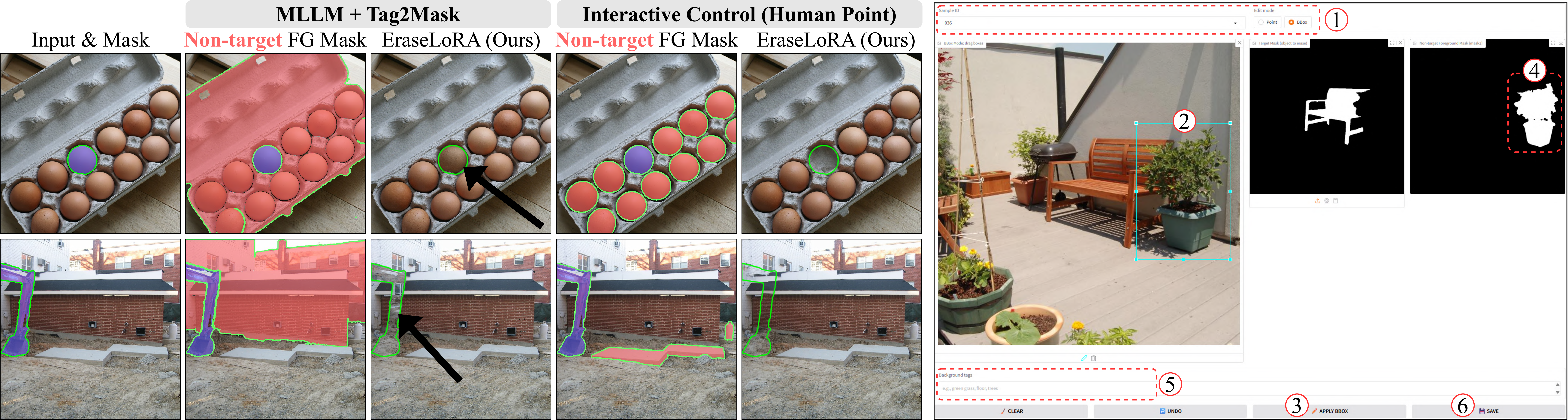} 
    \caption{\textbf{Interactive Control.} (Right) Our interactive interface allows users to generate customized non-target foreground masks for BFE (\cref{sec:bfe}) based on manual points or bounding boxes and background tags. (Left) These human-guided non-target foreground exclusion and background tag selection effectively correct challenging failure cases of EraseLoRA.}
    \label{fig:appendix_human_points}
\end{figure}

\textbf{Interactive Control.}
While EraseLoRA effectively removes the target automatically via MLLM's background-aware reasoning~\cite{zhu2025internvl3}, we offer an interactive variant where users provide minimal guidance via an interactive interface to enhance user control and computational efficiency. 
This mode allows users to bypass MLLM overhead or correct potential imperfect predictions of its reasoning. To support these needs, our interactive variant follows a streamlined workflow (Fig.~\ref{fig:appendix_human_points}, right): (i) selecting a specific sample and the edit mode (\emph{e.g.}, Point or BBox prompts), (ii) providing visual guidance via points or bounding boxes on the input image, (iii) clicking the apply button to (iv) update and refine the precise non-target foreground mask by observing the extracted results, (v) generating descriptive background tags based on the input image, and (vi) clicking the save button to store the manual results for BRSA (\cref{sec:brsa}).

\textbf{Limitations.} While EraseLoRA achieves strong performance in object removal, several limitations remain. First, the use of large-scale MLLMs such as InternVL3-78B~\cite{zhu2025internvl3} and test-time optimization for fixed 500 iterations with LoRA adapters introduces additional computational overhead. Specifically, the full optimization process takes approximately three minutes per test image on the default backbone, SD3.5-M~\cite{esser2024dit}. Although no paired data are required, iterative optimization and MLLM queries increase latency and memory usage at inference time. However, this cost can be significantly mitigated by (i) employing lightweight MLLMs (\emph{e.g.}, 7B scale), which show comparable quality as shown in Tab.~\ref{tab:ab_robustness_mllms_tag2mask}, and (ii) applying an early stopping strategy. We observe that employing an early stopping strategy can reduce the average optimization cost to approximately 141 iterations, achieving a more than 3.5$\times$ speedup compared to the fixed 500-iterations baseline while preserving over 96.5\% of the background similarity (see Appendix \ref{subsec:limitations} for more details).
Second, the MLLM-guided background definition can be imperfect in complex scenes with subtle semantic boundaries or heavy occlusion, which may lead to incomplete or inaccurate removal. A more detailed analysis is provided in Appendix \ref{subsec:appendix_discussion}.

\section{Conclusion}
\label{sec:conclusion}

In this paper, we introduce EraseLoRA, a dataset-free object removal framework that leverages MLLM-guided background-aware reasoning and test-time adaptation to enable faithful background restoration. Building on the failure modes identified in existing dataset-free methods, EraseLoRA explicitly separates target, non-target foreground, and background, filters out distractors to prevent their regeneration, and integrates multiple background subtypes. As a plug-and-play and model-agnostic module, EraseLoRA consistently produces cleaner and more coherent background reconstructions across diffusion backbones and benchmarks without requiring any paired data. An interesting future direction is video object removal, where shared background context across frames can amortize the per-image optimization cost.

\section*{Acknowledgements}
This work was partly supported by the KHIDI grant funded by the Korean government (MOHW) [No.RS-2025-02307233], the NRF or IITP grants funded by the Korean government (MSIT) [No.RS-2026-25472075, No.RS-2025-02305581, No.RS-2025-25442338 (AI Star Fellowship-SNU), and No.RS-2021II211343 (SNU AI)], the ITIP grant funded by the Korean government (MOTIR) [No.RS-2026-25549946], the Advanced GPU Utilization and AI Computing Infrastructure Enhancement User Support Programs funded by the Korean government (MSIT) [No.05-26-04-0094], the Research grant from SNU, and the Strategic Hub grant for International Research Collaboration of SNU. Kyungsu Kim is affiliated with the School of Transdisciplinary Innovations, Department of Biomedical Science, Interdisciplinary Program in Artificial Intelligence (IPAI), Medical Research Center, and AI Institute at SNU.

\clearpage
\bibliographystyle{splncs04}
\bibliography{main}

@String(IJCV  = {Int. J. Comput. Vis.})

@String(CVPR  = {IEEE Conf. Comput. Vis. Pattern Recog.})

@String(ICCV  = {Int. Conf. Comput. Vis.})

@String(ECCV  = {Eur. Conf. Comput. Vis.})

@String(NeurIPS = {Adv. Neural Inform. Process. Syst.})

@String(ICML  = {Int. Conf. Mach. Learn.})

@String(ICLR  = {Int. Conf. Learn. Represent.})

@String(BMVC  = {Brit. Mach. Vis. Conf.})

@String(AAAI  = {AAAI})

@String(TIP   = {IEEE Trans. Image Process.})

@String(WACV = {IEEE/CVF Winter Conf. Appl. Comput. Vis.})

@String(IJCV  = {IJCV})

@String(CVPR  = {CVPR})

@String(ICCV  = {ICCV})

@String(ECCV  = {ECCV})

@String(NeurIPS = {NeurIPS})

@String(ICML  = {ICML})

@String(ICLR  = {ICLR})

@String(BMVC  =	{BMVC})

@String(TIP   = {IEEE TIP})

@String(WACV = {IEEE WACV})

@inproceedings{Goodfellow2014gan,
 title = {Generative Adversarial Nets},
 author = {Goodfellow, Ian J. and Pouget-Abadie, Jean and Mirza, Mehdi and Xu, Bing and Warde-Farley, David and Ozair, Sherjil and Courville, Aaron and Bengio, Yoshua},
 booktitle = NeurIPS,
 volume = {27},
 year = {2014}
}

@inproceedings{ho2020ddpm,
 title = {Denoising Diffusion Probabilistic Models},
 author = {Ho, Jonathan and Jain, Ajay and Abbeel, Pieter},
 booktitle = NeurIPS,
 pages = {6840--6851},
 volume = {33},
 year = {2020}
}

@inproceedings{rombach2022sd,
  title     = {High-Resolution Image Synthesis With Latent Diffusion Models},
  author    = {Rombach, Robin and Blattmann, Andreas and Lorenz, Dominik and Esser, Patrick and Ommer, Bj\"orn},
  booktitle= CVPR,
  pages     = {10684--10695},
  year      = {2022}
}

@inproceedings{podell2024sdxl,
 title = {{SDXL}: Improving Latent Diffusion Models for High-Resolution Image Synthesis},
 author = {Podell, Dustin and English, Zion and Lacey, Kyle and Blattmann, Andreas and Dockhorn, Tim and M\"{u}ller, Jonas and Penna, Joe and Rombach, Robin},
 booktitle = ICLR,
 pages = {1862--1874},
 year = {2024}
}

@inproceedings{esser2024dit,
  title = 	 {Scaling Rectified Flow Transformers for High-Resolution Image Synthesis},
  author =       {Esser, Patrick and Kulal, Sumith and Blattmann, Andreas and Entezari, Rahim and M\"{u}ller, Jonas and Saini, Harry and Levi, Yam and Lorenz, Dominik and Sauer, Axel and Boesel, Frederic and Podell, Dustin and Dockhorn, Tim and English, Zion and Rombach, Robin},
  booktitle =  ICML,
  pages = 	 {12606--12633},
  volume = 	 {235},
  year = 	 {2024},
}

@inproceedings{Xie2023smartbrush,
  title     = {{SmartBrush}: Text and Shape Guided Object Inpainting With Diffusion Model},
  author    = {Xie, Shaoan and Zhang, Zhifei and Lin, Zhe and Hinz, Tobias and Zhang, Kun},
  booktitle = CVPR,
  pages     = {22428--22437},
  year      = {2023},
}

@inproceedings{zhuang2024powerpaint,
  title={A task is worth one word: Learning with task prompts for high-quality versatile image inpainting},
  author={Zhuang, Junhao and Zeng, Yanhong and Liu, Wenran and Yuan, Chun and Chen, Kai},
  booktitle=ECCV,
  pages={195--211},
  year={2024},
}

@inproceedings{liu2025erasediff,
  title     = {{Erase Diffusion}: Empowering Object Removal Through Calibrating Diffusion Pathways},
  author    = {Liu, Yi and Zhou, Hao and Cui, Benlei and Shang, Wenxiang and Lin, Ran},
  booktitle = CVPR,
  pages     = {2418--2427},
  year      = {2025},
}

@inproceedings{jiang2025smarteraser,
  title     = {{SmartEraser}: Remove Anything from Images using Masked-Region Guidance},
  author    = {Jiang, Longtao and Wang, Zhendong and Bao, Jianmin and Zhou, Wengang and Chen, Dongdong and Shi, Lei and Chen, Dong and Li, Houqiang},
  booktitle = CVPR,
  pages     = {24452--24462},
  year      = {2025},
}

@inproceedings{ekin2024clipaway,
 title = {{CLIPAway}: Harmonizing focused embeddings for removing objects via diffusion models},
 author = {Ekin, Yi\u{g}it and Yildirim, Ahmet Burak and Caglar, Erdem Eren and Erdem, Aykut and Erdem, Erkut and Dundar, Aysegul},
 booktitle = NeurIPS,
 volume = {37},
 pages = {17572--17601},
 year = {2024},
 doi = {10.52202/079017-0559},
}

@inproceedings{jia2025designedit,
  title = {{DesignEdit}: Unify Spatial-Aware Image Editing via Training-free Inpainting with a Multi-Layered Latent Diffusion Framework},
  author = {Jia, Yueru and Cheng, Aosong and Yuan, Yuhui and Wang, Chuke and Li, Ji and Jia, Huizhu and Zhang, Shanghang},
  booktitle = AAAI,
  volume={39},
  pages={3958--3966},
  year={2025},
  doi = {10.1609/aaai.v39i4.32414},
}

@inproceedings{sun2025attentiveeraser,
  title={{Attentive Eraser}: Unleashing Diffusion Model's Object Removal Potential via Self-Attention Redirection Guidance},
  author={Sun, Wenhao and Dong, Xue-Mei and Cui, Benlei and Tang, Jingqun},
  booktitle=AAAI,
  volume={39},
  pages={20734--20742},
  year={2025},
  doi = {10.1609/aaai.v39i19.34285},
}

@inproceedings{zhao2021comodgan,
  title={Large Scale Image Completion via Co-Modulated Generative Adversarial Networks},
  author={Zhao, Shengyu and Cui, Jonathan and Sheng, Yilun and Dong, Yue and Liang, Xiao and Chang, Eric I and Xu, Yan},
  booktitle=ICLR,
  year={2021}
}

@inproceedings{zuo2023ganseg,
  title={Generative Image Inpainting with Segmentation Confusion Adversarial Training and Contrastive Learning},
  author = {Zuo, Zhiwen and Zhao, Lei and Li, Ailin and Wang, Zhizhong and Zhang, Zhanjie and Chen, Jiafu and Xing, Wei and Lu, Dongming},
  booktitle=AAAI,
  volume={37},
  pages={3888--3896},
  year = {2023},
  doi = {10.1609/aaai.v37i3.25502},
}

@inproceedings{sargsyan2023migan,
  title     = {{MI-GAN}: A Simple Baseline for Image Inpainting on Mobile Devices},
  author    = {Sargsyan, Andranik and Navasardyan, Shant and Xu, Xingqian and Shi, Humphrey},
  booktitle = ICCV,
  pages     = {7335--7345},
  year      = {2023},
}

@inproceedings{yang2023paintbyexample,
  title     = {{Paint by Example}: Exemplar-Based Image Editing With Diffusion Models},
  author    = {Yang, Binxin and Gu, Shuyang and Zhang, Bo and Zhang, Ting and Chen, Xuejin and Sun, Xiaoyan and Chen, Dong and Wen, Fang},
  booktitle = CVPR,
  pages     = {18381--18391},
  year      = {2023},
}

@inproceedings{manukyan2025hdpainter,
  title = {{HD-Painter}: High-Resolution and Prompt-Faithful Text-Guided Image Inpainting with Diffusion Models},
  author = {Manukyan, Hayk and Sargsyan, Andranik and Atanyan, Barsegh and Wang, Zhangyang and Navasardyan, Shant and Shi, Humphrey},
  booktitle = ICLR,
  pages = {96301--96330},
  year = {2025},
}

@inproceedings{tianyidan2025anywhere,
  title={{Anywhere}: A Multi-Agent Framework for User-Guided, Reliable, and Diverse Foreground-Conditioned Image Generation},
  author = {Xie, Tianyidan and Ma, Rui and Wang, Qian and Ye, Xiaoqian and Liu, Feixuan and Tai, Ying and Zhang, Zhenyu and Wang, Lanjun and Yi, Zili},
  booktitle=AAAI,
  volume={39},
  pages={7410--7418},
  year = {2025},
  doi = {10.1609/aaai.v39i7.32797},
}

@inproceedings{fanelli2025idreammypainting,
  title={{I Dream My Painting}: Connecting MLLMs and Diffusion Models via Prompt Generation for Text-Guided Multi-Mask Inpainting},
  author={Fanelli, Nicola and Vessio, Gennaro and Castellano, Giovanna},
  booktitle=WACV,
  pages={6073--6082},
  year={2025},
}

@inproceedings{zhou2025fireedit,
  title     = {{FireEdit}: Fine-grained Instruction-based Image Editing via Region-aware Vision Language Model},
  author    = {Zhou, Jun and Li, Jiahao and Xu, Zunnan and Li, Hanhui and Cheng, Yiji and Hong, Fa-Ting and Lin, Qin and Lu, Qinglin and Liang, Xiaodan},
  booktitle = CVPR,
  pages     = {13093--13103},
  year      = {2025},
}

@inproceedings{zhu2025entityerasure,
  title     = {{EntityErasure}: Erasing Entity Cleanly via Amodal Entity Segmentation and Completion},
  author    = {Zhu, Yixing and Zhang, Qing and Wang, Yitong and Nie, Yongwei and Zheng, Wei-Shi},
  booktitle = CVPR,
  pages     = {28274--28283},
  year      = {2025},
}

@inproceedings{liu2024groundingdino,
  title = {{Grounding DINO}: Marrying {DINO} with Grounded Pre-training for Open-Set Object Detection},
  author = {Liu, Shilong and Zeng, Zhaoyang and Ren, Tianhe and Li, Feng and Zhang, Hao and Yang, Jie and Jiang, Qing and Li, Chunyuan and Yang, Jianwei and Su, Hang and Zhu, Jun and Zhang, Lei}, 
  booktitle=ECCV,
  pages={38--55},
  year={2024},
  doi = {10.1007/978-3-031-72970-6_3},
}

@inproceedings{ravi2025sam2,
  title={{SAM} 2: Segment Anything in Images and Videos},
  author={Ravi, Nikhila and Gabeur, Valentin and Hu, Yuan-Ting and Hu, Ronghang and Ryali, Chaitanya and Ma, Tengyu and Khedr, Haitham and R{\"a}dle, Roman and Rolland, Chloe and Gustafson, Laura and Mintun, Eric and Pan, Junting and Alwala, Kalyan Vasudev and Carion, Nicolas and Wu, Chao-Yuan and Girshick, Ross and Doll{\'a}r, Piotr and Feichtenhofer, Christoph},
  booktitle = ICLR,
  year = {2025},
}

@misc{flux2024,
    title={FLUX},
    author={{Black Forest Labs}},
    year={2024},
    howpublished={\url{https://github.com/black-forest-labs/flux}},
    note={Accessed 2026-06-26}
}

@article{wei2025omnieraser,
  title={{OmniEraser}: Remove Objects and Their Effects in Images with Paired Video-Frame Data},
  author={Wei, Runpu and Yin, Zijin and Zhang, Shuo and Zhou, Lanxiang and Wang, Xueyi and Ban, Chao and Cao, Tianwei and Sun, Hao and He, Zhongjiang and Liang, Kongming and Ma, Zhanyu},
  journal={arXiv preprint arXiv:2501.07397},
  year={2025}
}

@inproceedings{caron2021dino,
  title     = {Emerging Properties in Self-Supervised Vision Transformers},
  author    = {Caron, Mathilde and Touvron, Hugo and Misra, Ishan and J\'egou, Herv\'e and Mairal, Julien and Bojanowski, Piotr and Joulin, Armand},
  booktitle = ICCV,
  pages     = {9650--9660},
  year      = {2021},
}

@article{zhu2025internvl3,
  title={{InternVL3}: Exploring Advanced Training and Test-Time Recipes for Open-Source Multimodal Models},
  author={Zhu, Jinguo and Wang, Weiyun and Chen, Zhe and Liu, Zhaoyang and Ye, Shenglong and Gu, Lixin and Tian, Hao and Duan, Yuchen and Su, Weijie and Shao, Jie and others},
  journal={arXiv preprint arXiv:2504.10479},
  year={2025}
}

@article{bai2025qwen2.5vl,
  title={{Qwen2.5-VL} Technical Report},
  author={Bai, Shuai and Chen, Keqin and Liu, Xuejing and Wang, Jialin and Ge, Wenbin and Song, Sibo and Dang, Kai and Wang, Peng and Wang, Shijie and Tang, Jun and others},
  journal={arXiv preprint arXiv:2502.13923},
  year={2025}
}

@article{bai2025qwen3,
  title={{Qwen3-VL} Technical Report},
  author={Bai, Shuai and Cai, Yuxuan and Chen, Ruizhe and Chen, Keqin and Chen, Xionghui and Cheng, Zesen and Deng, Lianghao and Ding, Wei and Gao, Chang and Ge, Chunjiang and others},
  journal={arXiv preprint arXiv:2511.21631},
  year={2025}
}

@article{ren2024groundedsam,
  title={{Grounded SAM}: Assembling Open-World Models for Diverse Visual Tasks},
  author={Ren, Tianhe and Liu, Shilong and Zeng, Ailing and Lin, Jing and Li, Kunchang and Cao, He and Chen, Jiayu and Huang, Xinyu and Chen, Yukang and Yan, Feng and Zeng, Zhaoyang and Zhang, Hao and Li, Feng and Yang, Jie and Li, Hongyang and Jiang, Qing and Zhang, Lei},
  journal={arXiv preprint arXiv:2401.14159},
  year={2024}
}

@article{kuznetsova2020openimagesv7,
  title   = {{The Open Images Dataset V4}: Unified Image Classification, Object Detection, and Visual Relationship Detection at Scale},
  author  = {Kuznetsova, Alina and Rom, Hassan and Alldrin, Neil and Uijlings, Jasper and Krasin, Ivan and Pont-Tuset, Jordi and Kamali, Shahab and Popov, Stefan and Malloci, Matteo and Kolesnikov, Alexander and Duerig, Tom and Ferrari, Vittorio},
  journal = IJCV,
  volume  = {128},
  pages   = {1956--1981},
  year    = {2020},
  doi     = {10.1007/s11263-020-01316-z},
}

@inproceedings{sagong2022rord,
  author    = {Sagong, Min-Cheol and Yeo, Yoon-Jae and Jung, Seung-Won and Ko, Sung-Jea},
  title     = {{RORD}: A Real-world Object Removal Dataset},
  booktitle = BMVC,
  pages     = {542},
  year      = {2022}
}

@inproceedings{liu2023llava,
 title = {Visual Instruction Tuning},
 author = {Liu, Haotian and Li, Chunyuan and Wu, Qingyang and Lee, Yong Jae},
 booktitle = NeurIPS,
 pages = {34892--34916},
 volume = {36},
 year = {2023},
}

@inproceedings{wang2021tta,
  title={Tent: Fully Test-Time Adaptation by Entropy Minimization},
  author={Wang, Dequan and Shelhamer, Evan and Liu, Shaoteng and Olshausen, Bruno and Darrell, Trevor},
  booktitle=ICLR,
  year={2021},
}

@inproceedings{kim2025chainofzoom,
  title = {{Chain-of-Zoom}: Extreme Super-Resolution via Scale Autoregression and Preference Alignment},
  author={Kim, Bryan Sangwoo and Kim, Jeongsol and Ye, Jong Chul},
  booktitle = NeurIPS,
  volume    = {38},
  year = {2025},
}

@inproceedings{chen2024freecompose,
  title = {{FreeCompose}: Generic Zero-Shot Image Composition with Diffusion Prior},
  author={Chen, Zhekai and Wang, Wen and Yang, Zhen and Yuan, Zeqing and Chen, Hao and Shen, Chunhua},
  booktitle = ECCV,
  pages = {70--87},
  year = {2024},
  doi = {10.1007/978-3-031-72643-9_5},
}

@inproceedings{wang2024genartist,
 title = {{GenArtist}: Multimodal LLM as an Agent for Unified Image Generation and Editing},
 author={Wang, Zhenyu and Li, Aoxue and Li, Zhenguo and Liu, Xihui},
 booktitle = NeurIPS,
 volume={37},
 pages={128374--128395},
 year = {2024},
}

@inproceedings{Qu2025silmm,
  title     = {{SILMM}: Self-Improving Large Multimodal Models for Compositional Text-to-Image Generation},
  author    = {Qu, Leigang and Li, Haochuan and Wang, Wenjie and Liu, Xiang and Li, Juncheng and Nie, Liqiang and Chua, Tat-Seng},
  booktitle =  CVPR,
  pages     = {18497--18508},
  year      = {2025},
}

@inproceedings{zhang2023magicbrush,
  title = {{MagicBrush}: a manually annotated dataset for instruction-guided image editing},
  author={Zhang, Kai and Mo, Lingbo and Chen, Wenhu and Sun, Huan and Su, Yu},
  booktitle = NeurIPS,
  volume    = {36},
  pages={31428--31449},
  year = {2023},
}

@inproceedings{li2024tuning,
  title = {Tuning-Free Image Customization with Image and Text Guidance},
  author={Li, Pengzhi and Nie, Qiang and Chen, Ying and Jiang, Xi and Wu, Kai and Lin, Yuhuan and Liu, Yong and Peng, Jinlong and Wang, Chengjie and Zheng, Feng},
  booktitle = ECCV,
  pages = {233--250},
  year = {2024},
  doi = {10.1007/978-3-031-73116-7_14},
}

@inproceedings{hu2022lora,
  title={{LoRA}: Low-Rank Adaptation of Large Language Models},
  author={Hu, Edward J. and Shen, Yelong and Wallis, Phillip and Allen-Zhu, Zeyuan and Li, Yuanzhi and Wang, Shean and Wang, Lu and Chen, Weizhu},
  booktitle=ICLR,
  year={2022},
}

@inproceedings{suvorov2022resolution,
  title     = {Resolution-Robust Large Mask Inpainting With Fourier Convolutions},
  author    = {Suvorov, Roman and Logacheva, Elizaveta and Mashikhin, Anton and Remizova, Anastasia and Ashukha, Arsenii and Silvestrov, Aleksei and Kong, Naejin and Goka, Harshith and Park, Kiwoong and Lempitsky, Victor},
  booktitle = WACV,
  pages     = {2149--2159},
  year      = {2022},
}

@inproceedings{li2022mat,
  title     = {{MAT}: Mask-Aware Transformer for Large Hole Image Inpainting},
  author    = {Li, Wenbo and Lin, Zhe and Zhou, Kun and Qi, Lu and Wang, Yi and Jia, Jiaya},
  booktitle =  CVPR,
  pages     = {10758--10768},
  year      = {2022},
}

@article{wang2004ssim,
  title={Image quality assessment: from error visibility to structural similarity}, 
  author={Wang, Zhou and Bovik, Alan C and Sheikh, Hamid R and Simoncelli, Eero P},
  journal=TIP,  
  volume={13},
  number={4},
  pages={600--612},
  year={2004},
  doi={10.1109/TIP.2003.819861},
}

@inproceedings{zhao2025marine,
  title={Mitigating Object Hallucination in Large Vision-Language Models via Image-Grounded Guidance},
  author={Zhao, Linxi and Deng, Yihe and Zhang, Weitong and Gu, Quanquan},
  booktitle=ICML,
  volume={267},
  pages={77461--77486},
  year={2025},
}

@inproceedings{kim2025early,
  title     = {Early Timestep Zero-Shot Candidate Selection for Instruction-Guided Image Editing},
  author    = {Kim, Joowon and Lee, Ziseok and Cho, Donghyeon and Jo, Sanghyun and Jung, Yeonsung and Kim, Kyungsu and Yang, Eunho},
  booktitle = ICCV,
  pages     = {18844--18854},
  year      = {2025},
}

@inproceedings{zhu2025kv,
  title     = {{KV-Edit}: Training-Free Image Editing for Precise Background Preservation},
  author    = {Zhu, Tianrui and Zhang, Shiyi and Shao, Jiawei and Tang, Yansong},
  booktitle = ICCV,
  pages     = {16607--16617},
  year      = {2025},
}

@article{simeoni2025dinov3,
  title={DINOv3},
  author={Sim{\'e}oni, Oriane and Vo, Huy V. and Seitzer, Maximilian and Baldassarre, Federico and Oquab, Maxime and Jose, Cijo and Khalidov, Vasil and Szafraniec, Marc and Yi, Seungeun and Ramamonjisoa, Micha{\"e}l and Massa, Francisco and Haziza, Daniel and Wehrstedt, Luca and Wang, Jianyuan and Darcet, Timoth{\'e}e and Moutakanni, Th{\'e}o and Sentana, Leonel and Roberts, Claire and Vedaldi, Andrea and Tolan, Jamie and Brandt, John and Couprie, Camille and Mairal, Julien and J{\'e}gou, Herv{\'e} and Labatut, Patrick and Bojanowski, Piotr},
  journal={arXiv preprint arXiv:2508.10104},
  year={2025}
}

@inproceedings{yu2025omnipaint,
  title     = {{OmniPaint}: Mastering Object-Oriented Editing via Disentangled Insertion-Removal Inpainting},
  author    = {Yu, Yongsheng and Zeng, Ziyun and Zheng, Haitian and Luo, Jiebo},
  booktitle = ICCV,
  pages     = {17324--17334},
  year      = {2025},
}

@inproceedings{zhang2018lpips,
  title = {The Unreasonable Effectiveness of Deep Features as a Perceptual Metric},
  author = {Zhang, Richard and Isola, Phillip and Efros, Alexei A. and Shechtman, Eli and Wang, Oliver},
  booktitle = CVPR,
  pages={586--595},
  year = {2018}
}

@InProceedings{wang2025yoloe,
  title     = {{YOLOE}: Real-Time Seeing Anything},
  author    = {Wang, Ao and Liu, Lihao and Chen, Hui and Lin, Zijia and Han, Jungong and Ding, Guiguang},
  booktitle = ICCV,
  pages     = {24591--24602},
  year      = {2025},
}

@inproceedings{kim2025seg4diff,
  title = {{Seg4Diff}: Unveiling Open-Vocabulary Segmentation in Text-to-Image Diffusion Transformers},
  author = {Kim, Chaehyun and Shin, Heeseong and Hong, Eunbeen and Yoon, Heeji and Arnab, Anurag and Seo, Paul Hongsuck and Hong, Sunghwan and Kim, Seungryong},
  booktitle = NeurIPS,
  volume    = {38},
  year = {2025},
}

@inproceedings{carion2025sam3,
  title     = {{SAM 3}: Segment Anything with Concepts},
  author    = {Carion, Nicolas and Gustafson, Laura and Hu, Yuan-Ting and Debnath, Shoubhik and Hu, Ronghang and Coll-Vinent, Didac Suris and Ryali, Chaitanya and Alwala, Kalyan Vasudev and Khedr, Haitham and Huang, Andrew and Lei, Jie and Ma, Tengyu and Guo, Baishan and Kalla, Arpit and Marks, Markus and Greer, Joseph and Wang, Meng and Sun, Peize and R{\"a}dle, Roman and Afouras, Triantafyllos and Mavroudi, Effrosyni and Xu, Katherine and Wu, Tsung-Han and Zhou, Yu and Momeni, Liliane and Hazra, Rishi and Ding, Shuangrui and Vaze, Sagar and Porcher, Francois and Li, Feng and Li, Siyuan and Kamath, Aishwarya and Cheng, Ho Kei and Dollar, Piotr and Ravi, Nikhila and Saenko, Kate and Zhang, Pengchuan and Feichtenhofer, Christoph},
  booktitle = ICLR,
  year      = {2026},
}

@inproceedings{fu2026EffectErase,
  title     = {{EffectErase}: Joint Video Object Removal and Insertion for High-Quality Effect Erasing},
  author    = {Fu, Yang and Zheng, Yike and Dai, Ziyun and Ding, Henghui},
  booktitle = CVPR,
  pages     = {2005--2014},
  year      = {2026},
}

@inproceedings{zhao2026objectclear,
  title     = {Precise Object and Effect Removal with Adaptive Target-Aware Attention},
  author    = {Zhao, Jixin and Wang, Zhouxia and Yang, Peiqing and Zhou, Shangchen},
  booktitle = CVPR,
  pages     = {19370-19379},
  year      = {2026},
}

% ---- Supplementary ----
%
\clearpage
\appendix

\section{Preliminaries} \label{sec:preliminaries}
\textbf{Diffusion Models.} Diffusion models generate images by learning a reverse denoising process that gradually transforms noise into image \cite{podell2024sdxl, esser2024dit}. In a typical text-to-image implementation, an input image $I$ is first mapped into a latent representation $z_0 = \text{Enc}(I)$ through a VAE encoder. A denoising network $\epsilon_\theta$ then iteratively predicts and removes the noise from $z_t$, optionally conditioned on a text embedding $c$. Finally, the denoised latent $\hat{z}_0$ is decoded back into the image space by a VAE decoder, yielding $\hat{I} = \text{Dec}(\hat{z}_0)$. This framework allows controllable generation by conditioning on text prompts or other external signals. 

\textbf{Attention Mechanisms in Diffusion Models.} In latent diffusion models \cite{rombach2022sd}, attention blocks regulate how information flows during denoising. There are two types of attention: self-attention captures dependencies among latent tokens, while cross-attention establishes interactions between latent tokens and external condition tokens such as text prompts. For a latent $z_t \in \mathbb{R}^{HW \times d}$ at step $t$, self-attention computes a weight matrix $A^{\text{self}} = \mathrm{softmax}(QK^\top/\sqrt{d}) \in [0,1]^{HW \times HW}$, where each row sums to $1$, which represents how strongly one latent token attends to all other tokens. Cross-attention follows the same principle but aligns latent queries with condition tokens $C \in \mathbb{R}^{L \times d}$, yielding $A^{\text{cross}} \in [0,1]^{HW \times L}$. Here each row of $A^{\text{cross}}$ sums to $1$, corresponding to $\sum_{i=1}^L A^{\text{cross}}_i[p] = 1$ for each spatial index $p$, where $A^{\text{cross}}_i \in [0,1]^{HW}$ denotes the cross-attention matrix for condition token $i$. This indicates how strongly each latent token relates to the $L$ condition tokens. In practice, the way attention is computed has evolved with the design of 
modern diffusion backbones. Earlier text-to-image diffusion models \cite{rombach2022sd, podell2024sdxl} compute attention directly at the latent level on feature maps. However, recent powerful text-to-image diffusion models \cite{esser2024dit, flux2024} group multiple latent tokens into larger patch tokens (\emph{latent patchify}) and compute attention at the patch level to enable more efficient and scalable processing of high-resolution images.

\section{Details of EraseLoRA} \label{sec:details_of_method}

\subsection{Implementation Details} \label{subsec:implementation_details}
For reproducibility, we provide the implementation details and hyperparameters of EraseLoRA. During test-time adaptation, only the LoRA adapters \cite{hu2022lora} inserted into the attention blocks are updated, while all backbone parameters remain frozen. 
Optimization is performed using the test-time adaptation objective defined in \cref{sec:brsa}. All baseline methods \cite{esser2024dit, sun2025attentiveeraser, jia2025designedit, podell2024sdxl, flux2024, zhuang2024powerpaint, ekin2024clipaway, jiang2025smarteraser, wei2025omnieraser, zhu2025entityerasure} are reproduced using official implementations when available, or re-implemented following the descriptions in their respective papers.

\textbf{Dice Score.} 
We use a soft Dice coefficient between a continuous attention map $X$ and a binary region indicator $Y$:
\[
\mathrm{Dice}(X, Y) = 
\frac{2 \sum_p X[p]Y[p]}{\sum_p X[p] + \sum_p Y[p] + \epsilon},
\]
where $\epsilon$ is a small constant added for numerical stability.

\textbf{Normalized Cross-Attention.}
For each background subtype tag $b \in \mathcal{B}$, we compute a raw spatial attention map $\tilde{A}_b$ from the cross-attention values of the diffusion model. Let $\mathcal{T}_b$ denote the set of text token indices corresponding to tag $b$. For each spatial index $p$, we average the cross-attention responses over attention layers, heads, and the text tokens in $\mathcal{T}_b$:
\[
\tilde{A}_b[p]
=
\frac{1}{|\mathcal{L}|\,|\mathcal{H}|\,|\mathcal{T}_b|}
\sum_{\ell \in \mathcal{L}}
\sum_{h \in \mathcal{H}}
\sum_{i \in \mathcal{T}_b}
A_{\ell,h}^{\mathrm{cross}}[p,i],
\]
where $\mathcal{L}$ and $\mathcal{H}$ denote the sets of attention layers and heads, respectively, and $A_{\ell,h}^{\mathrm{cross}}[p,i]$ denotes the cross-attention value between spatial index $p$ and text token $i$.

We then apply tag-wise normalization across background subtype tags:
\[
A_b[p] =
\frac{
\exp(\tau \tilde{A}_b[p])
}{
\sum_{b' \in \mathcal{B}} \exp(\tau \tilde{A}_{b'}[p])
},
\]
where $\tau$ controls the sharpness of subtype assignment, and we set $\tau = 100$ in all experiments.

Finally, the dominant subtype aggregation map used in $\mathcal{L}_{\text{puzzle}}$ is defined as
\[
A^{\mathrm{dom}}[p] = \max_{b \in \mathcal{B}} A_b[p].
\]
This map records the strongest normalized response among background subtype tags at each spatial location. Matching $A^{\mathrm{dom}}$ to $\mathbf{1}_{M_T \cup M_B}$ promotes subtype aggregation on the target and clean background regions while suppressing responses on the non-target foreground mask $M_F$.

\begin{figure}[p]
    \centering
    \includegraphics[width=\linewidth]{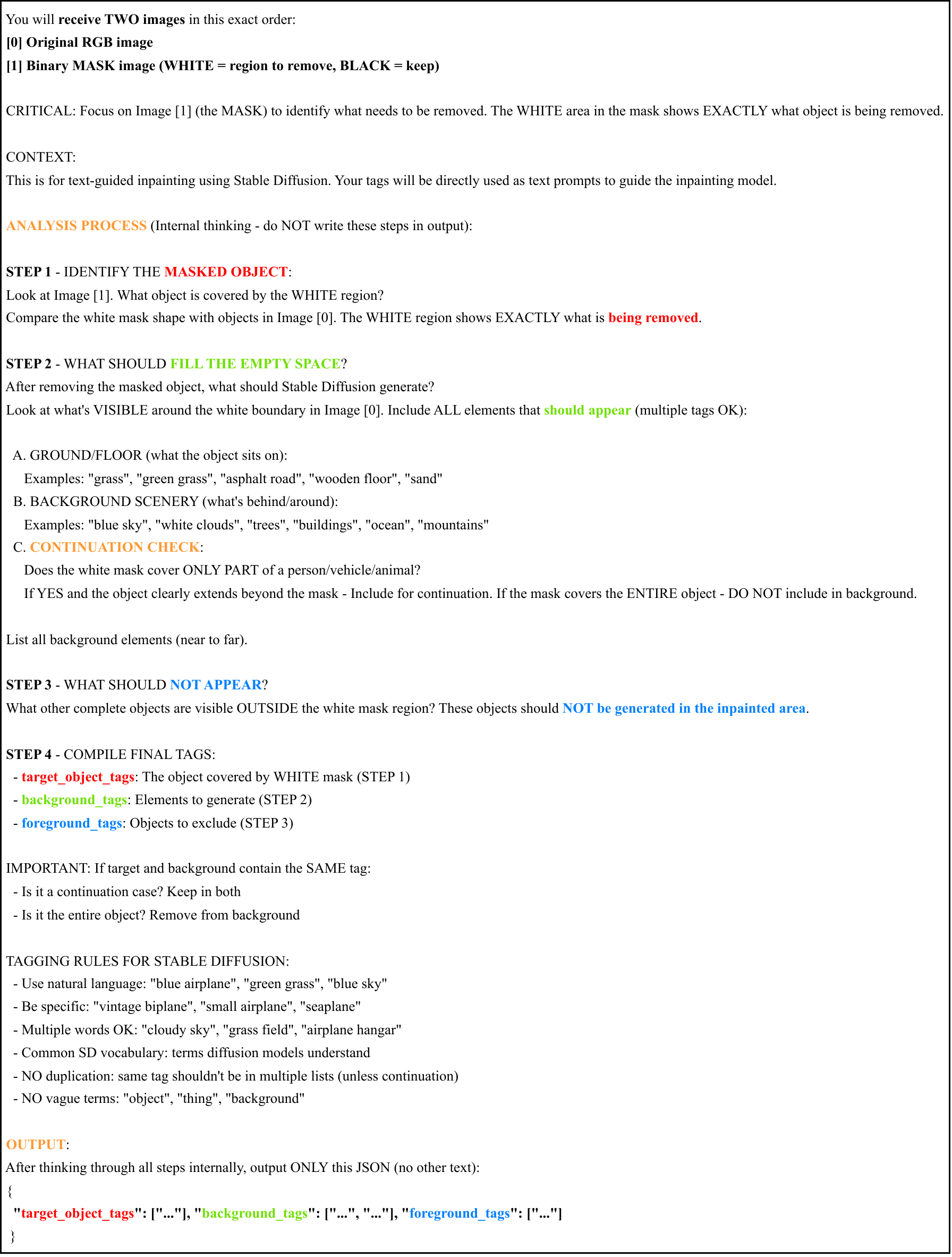} 
    \caption{Full MLLM prompt used in BFE (\cref{sec:bfe}) for background-aware reasoning and tag extraction from a single image–mask pair. The prompt guides a step-by-step reasoning that accounts for occlusions (orange line) to extract key cues: target objects (red line), background subtypes for reconstruction (green line), and non-target foregrounds to be excluded (blue line).}
    \label{fig:appendix_mllm_prompt}
\end{figure}

\subsubsection{Configurations.}

\begin{itemize}
    \item \textbf{Hardware.} All experiments are conducted on NVIDIA RTX PRO 6000 Blackwell Server Edition GPUs with 96GB VRAM using mixed-precision (FP16).
    \item \textbf{Diffusion Backbone choices.} SD3.5-M~\cite{esser2024dit} is used as the default text-to-image diffusion backbone based on its powerful text-image alignment ability. To validate flexibility across diverse modern diffusion backbones, we additionally test our method on several different diffusion architectures, including SD1.5~\cite{rombach2022sd}, SDXL~\cite{podell2024sdxl}, and FLUX.1~\cite{flux2024} (see Tab. \ref{tab:appendix_robustness_backbones}).
    \item \textbf{MLLM choices.} InternVL3-78B~\cite{zhu2025internvl3} is used in BFE (\cref{sec:bfe}) as the default MLLM due to its strong background-aware reasoning. To validate flexibility across different MLLMs, we additionally test representative models of various scales, including LLaVA-7B \cite{liu2023llava} and Qwen2.5-VL-72B \cite{bai2025qwen2.5vl}. We further include a hallucination-mitigated model, LLaVA-7B w/ MARINE \cite{zhao2025marine}, to examine the impact of object hallucination on removal quality (see Tab. \ref{tab:ab_robustness_mllms_tag2mask}, left). For all MLLM variants, we consistently apply the same full prompt as illustrated in Fig. \ref{fig:appendix_mllm_prompt} to ensure a fair comparison of their background-aware reasoning capabilities.
    {\item \textbf{Tag2Mask choices.} Grounding DINO~\cite{liu2024groundingdino} and SAM2~\cite{ravi2025sam2} are used in BFE (\cref{sec:bfe}) as the default Tag2Mask model to obtain pixel-level masks for MLLM-predicted tags. To validate flexibility across different Tag2Mask models, we additionally test three state-of-the-art segmentation models, including Seg4Diff~\cite{kim2025seg4diff}, YOLOE~\cite{wang2025yoloe} and SAM3~\cite{carion2025sam3} (see Tab. \ref{tab:ab_robustness_mllms_tag2mask}, right).}
    \item \textbf{LoRA details.} We test multiple ranks ($r\in\{16,32,64,128\}$) for the LoRA adapters inserted into attention blocks, and set the rank to $r=32$, providing the best results, for all experiments (see Fig. \ref{fig:appendix_tta_param}).
    \item \textbf{TTA iterations.} We compare different numbers of iterations ($\{100, \allowbreak 200, \allowbreak 300, \allowbreak 400, \allowbreak 500, \allowbreak 600\}$) during test-time adaptation. (see Fig. \ref{fig:appendix_tta_param}).
    While we set 500 iterations as the default for all experiments to ensure maximum quality, we also introduce an Early Stopping (E.S.) strategy to significantly improve efficiency while maintaining quality. We provide the technical details of this strategy in \cref{subsec:limitations}.
    \item \textbf{Loss weights.} The weight of the puzzle loss $\mathcal{L}_{\text{puzzle}}$ is set to $\lambda = 0.2$ in the TTA objective.
    \item \textbf{Computational cost of TTA.} The VRAM usage, number of additional parameters, and optimization time of BRSA (\cref{sec:brsa}) during test-time optimization are summarized in Tab. \ref{tab:appendix_robustness_backbones}. Only the LoRA parameters are updated during optimization and the updated LoRA weights are merged into the backbone afterward~\cite{hu2022lora}. As a result, EraseLoRA incurs no extra computational cost at inference time, consistent with standard text-to-image diffusion architectures~\cite{rombach2022sd, podell2024sdxl, esser2024dit, flux2024} (see Tab.~\ref{tab:appendix_removalbench_inference_cost}, right).
\end{itemize}

\subsubsection{Metric Details.}
To evaluate object-removal performance on unpaired datasets that do not have ground-truth images after removal, we use BG Sim. and FG Sim., two DINO-based cosine similarity metrics~\cite{caron2021dino}, along with BG Pres. (introduced in \cref{sec:experiment}).
Following the mask partition defined in Eq.~\ref{eq:region_label}, we use the target mask $M_T$, non-target foreground mask $M_F$, and background mask $M_B$ for metric computation.
We manually curate the corresponding masks to obtain reliable regions for evaluation and will release them publicly.

\textbf{BG Sim.}
Background Similarity (BG Sim.) measures how well the reconstructed masked
region aligns with the true background (higher is better). We compute it as the cosine similarity
between DINOv3 \cite{simeoni2025dinov3} features extracted from the background
region in the input image and the reconstructed region in the output:
\begin{equation}
\mathrm{BG\text{ }Sim.}
=
\frac{
f(I[M_B]) \cdot f(\hat{I}[M_T])
}{
\|f(I[M_B])\|\,\|f(\hat{I}[M_T])\|
}
\label{eq:bg_sim}
\end{equation}
where $M_B$ denotes the background mask and $M_T$ denotes the reconstructed target region.

\textbf{FG Sim.}
Foreground Similarity (FG Sim.) measures how much the reconstructed masked region
incorrectly refers to foreground content (lower is better). It is computed as the cosine similarity between DINOv3 \cite{simeoni2025dinov3} features extracted from the foreground region in the input image and the reconstructed region in the output. To discourage background-inconsistent restoration, we weight this score by
$(1 - \mathrm{BG\text{ }Sim.})$:
\begin{equation}
\mathrm{FG\text{ }Sim.}
=
(1 - \mathrm{BG\text{ }Sim.}) \cdot
\frac{
f(I[M_T \cup M_F]) \cdot f(\hat{I}[M_T])
}{
\|f(I[M_T \cup M_F])\|\,\|f(\hat{I}[M_T])\|
}
\label{eq:fg_sim}
\end{equation}
where $M_T \cup M_F$ denotes the all foreground region (target and non-target).

\textbf{BG Pres.} 
Background Preservation (BG Pres.) evaluates how well the unmasked region is maintained after object removal. It is computed as SSIM \cite{wang2004ssim} between the input image $I$ and reconstructed image $\hat{I}$ over the unmasked region $M_T^c$, following the protocol of recent editing works \cite{zhu2025kv, kim2025early}.

\begin{figure}[t]
    \centering
    \includegraphics[width=\linewidth]{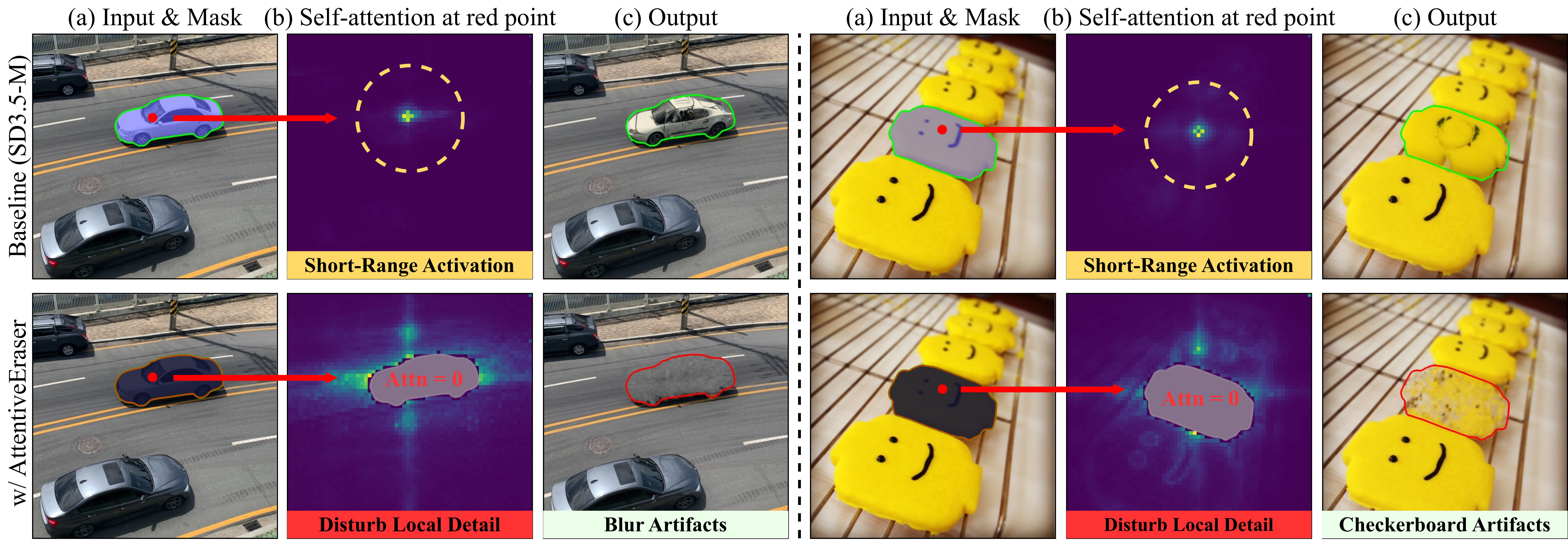} 
    \caption{\textbf{Artifacts from disrupting short-range self-attention.} Prior self-attention control methods \cite{jia2025designedit, sun2025attentiveeraser} suppress short-range activations (b), which erodes fine details and produces blur and checkerboard artifacts in the reconstructed background (c).}
    \label{fig:appendix_attention_artifacts}
\end{figure}

\subsection{Design Rationale} \label{subsec:attention_artifacts}
\subsubsection{Limitations of Attention Surgery.} 
Previous state-of-the-art dataset-free methods \cite{chen2024freecompose, jia2025designedit, sun2025attentiveeraser} redirect or block self-attention inside the mask so that the model focuses on unmasked context. Specifically, they block any interaction within the masked region itself by setting those attention values to zero (see Fig.~\ref{fig:appendix_attention_artifacts} (b)).
However, we identify two fundamental limitations in these methods. 

First, they are inherently unstable and often produce blur or structural artifacts (see Fig.~\ref{fig:appendix_attention_artifacts} (c)). They unintentionally disrupt short-range activations of self-attention, referring to the local interactions where latent tokens mainly attend to their nearby neighbors. These activations are crucial for preserving fine-grained details, and their disruption often leads to blurred textures. Furthermore, when applied to recent text-to-image diffusion models \cite{esser2024dit, flux2024} that compute attention at the patch level, blocking attention inside the mask amplifies instability, resulting in patch-wise artifacts such as checkerboard patterns. 
Second, they apply uniform attention constraints without distinguishing the diverse background subtypes in the unmasked region. By treating the entire context as a single and undifferentiated background, they indiscriminately attend to irrelevant features, such as using sidewalk details to reconstruct a road, and fail to preserve the unique structural priors of specific patterns such as road markings or cooling racks. This leads to textural blurring, structural misalignment, and unnatural boundaries between disparate subtypes (see Fig.~\ref{fig:artifact_attention} and Fig.~\ref{fig:appendix_attention_artifacts} (c)).

These failure modes stem from the absence of explicit background-aware reasoning, showing that prior self-attention control methods \cite{chen2024freecompose, jia2025designedit, sun2025attentiveeraser} fail to maintain fine-grained fidelity and produce visually inauthentic results, lacking robustness across modern text-to-image diffusion architectures \cite{esser2024dit, flux2024}.

\subsubsection{Effect of Test-Time Adaptation.}
Instead of explicitly blocking self-attention, BRSA (Sec.~\ref{sec:brsa}) provides a dataset-free mechanism for injecting BFE-inferred background cues into the diffusion model through test-time adaptation with LoRA adapters~\cite{hu2022lora}. Unlike dataset-driven approaches~\cite{zhuang2024powerpaint, jiang2025smarteraser, zhu2025entityerasure} that rely on paired before/after removal data or a dataset-level removal prior, BRSA adapts the frozen backbone to the current image context, allowing clean background information to be incorporated into the masked-region reconstruction through the adapted weights rather than explicit attention manipulation.

This test-time adaptation is a key design choice for inducing removal behavior without additional training data while avoiding artifacts caused by attention surgery. Together with the complementary BRSA objectives (Eqs.~\ref{eq:recon} and~\ref{eq:puzzle}), it enables background cues to be aggregated with textual and structural consistency. As a result, BRSA reconstructs the masked region using image-specific background information, reducing blur, checkerboard artifacts, and subtype-misaligned textures across diverse diffusion backbones \cite{rombach2022sd, podell2024sdxl, esser2024dit, flux2024}.

\section{Additional Quantitative Results} \label{sec:appendix_quantitative_results}
\subsection{Experimental Setup} \label{subsec:appendix_experiment_setup}
This section provides extended evaluation details on baselines, datasets, and metrics.

\textbf{Baselines.} Beyond the main comparison in Tab.~\ref{tab:quantitative_results}, we provide additional evaluations of FLUX.1-Fill~\cite{flux2024} to broaden the set of competitive baselines.

\begin{figure}[t]
    \centering
    \includegraphics[width=0.5\linewidth]{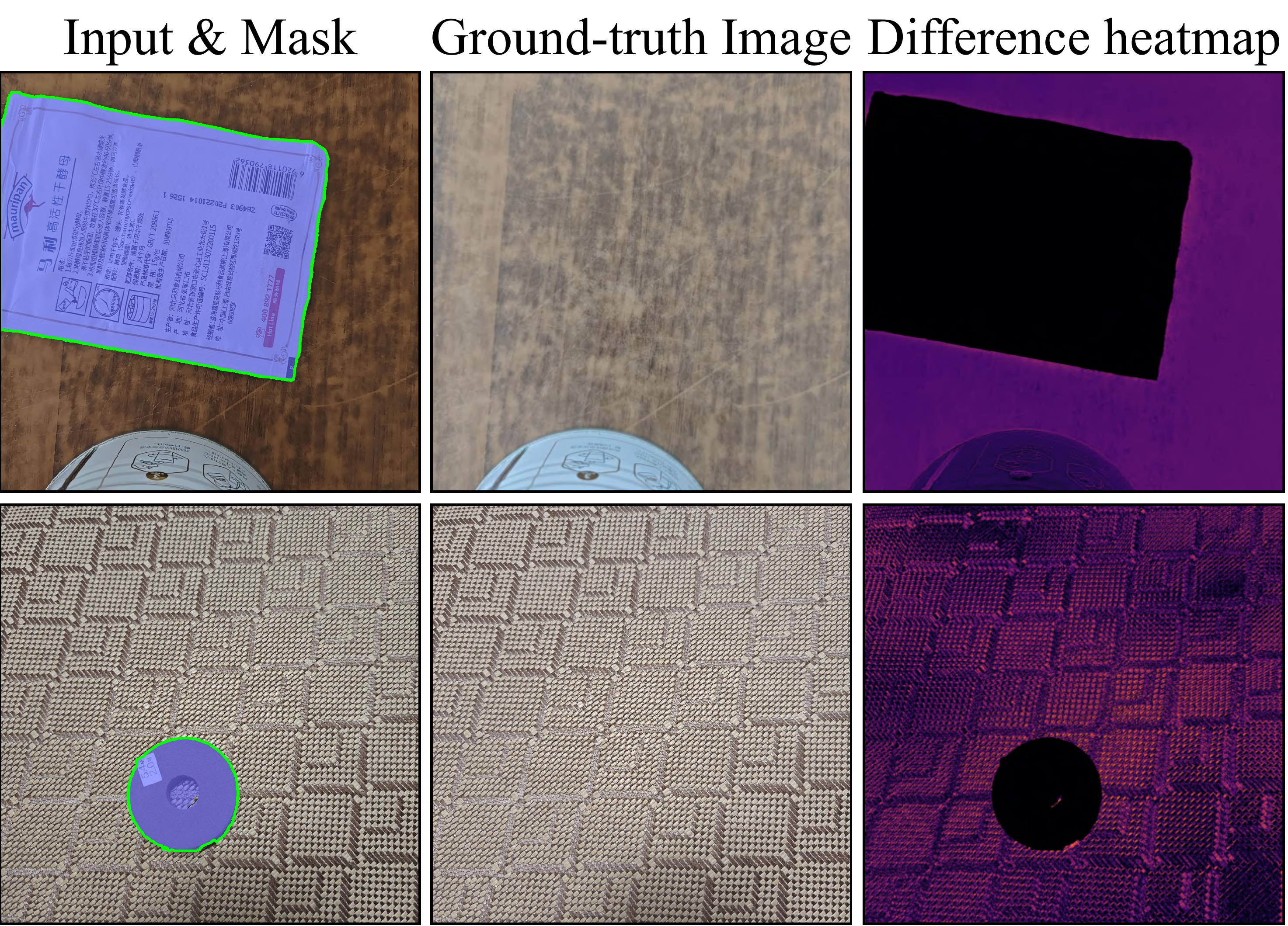} 
    \caption{\textbf{Color discrepancies on paired OmniPaint-Bench \cite{yu2025omnipaint} dataset.} Before/after object removal pairs show substantial color mismatch in target-unrelated regions, as shown in the difference heatmap.}
    \label{fig:appendix_ominpaintbench}
\end{figure}

\textbf{Benchmarks.} We further evaluate EraseLoRA on RemovalBench~\cite{wei2025omnieraser}, which provides 68 paired samples with ground-truth images after target removal. We also consider OmniPaint-Bench~\cite{yu2025omnipaint}, which offers 1,300 paired samples. However, we find that 1,000 samples overlap with RORD~\cite{sagong2022rord}, and the remaining 300 samples exhibit notable inconsistencies between the input and ground-truth images, such as mismatched colors in unmasked regions (see Fig.~\ref{fig:appendix_ominpaintbench}). 
Therefore, we exclude OmniPaint-Bench~\cite{yu2025omnipaint} from our evaluation, as its samples do not provide reliable ground-truth backgrounds for fair quantitative comparison.

\textbf{Extended metrics.} For unpaired object removal, we use three evaluation metrics: Foreground Similarity (FG Sim.), Background Similarity (BG Sim.), and Background Preservation (BG Pres.). For paired object removal, we further report representative fidelity metrics, PSNR, SSIM~\cite{wang2004ssim}, and LPIPS~\cite{zhang2018lpips}, to compare predictions against ground-truth backgrounds. Following the evaluation protocol of RemovalBench~\cite{wei2025omnieraser}, all paired metrics are computed on masked regions to measure pixel-level accuracy, structural consistency, and perceptual similarity (see Tab.~\ref{tab:appendix_removalbench_inference_cost}, left).

Although these evaluation metrics capture different aspects of image fidelity, they are still insufficient to determine whether object removal is semantically successful. In particular, they cannot verify whether the target object truly disappears without residual foreground traces, nor whether the reconstructed background is contextually plausible.

To address this, we follow recent VLM-based evaluation protocols \cite{sun2025attentiveeraser} for image inpainting and object removal and adapt them into a removal-specific metric, GPT-Metric, scored on a 0–100 scale.
GPT-Metric assesses object removal from a semantic perspective via two components: (1) a removal success rate, which checks whether the target object is correctly perceived as absent without any traces, and (2) a semantic perceptual score, which evaluates the quality of the reconstructed background, including contextual consistency and hallucination artifacts. Detailed quantitative results for GPT-Metric are reported in Tab.~\ref{tab:appendix_opiv7_rord_results}.

\subsection{Quantitative Analysis}
\textbf{Additional quantitative results.} 
Across OpenImages V7~\cite{kuznetsova2020openimagesv7}, RORD~\cite{sagong2022rord}, and RemovalBench~\cite{wei2025omnieraser}, EraseLoRA consistently surpasses all dataset-free approaches~\cite{esser2024dit, sun2025attentiveeraser, jia2025designedit} and remains competitive against dataset-driven methods~\cite{podell2024sdxl, flux2024, zhuang2024powerpaint, ekin2024clipaway, jiang2025smarteraser, zhu2025entityerasure, wei2025omnieraser}.
We observe that EraseLoRA improves background fidelity and alleviates regeneration of undesired foreground without any noise while preserving background.
Table~\ref{tab:appendix_opiv7_rord_results} and the left of Table~\ref{tab:appendix_removalbench_inference_cost} summarize the extended quantitative results across datasets.

\begin{table}[t]
    \centering
    \caption{Extended quantitative comparison with previous state-of-the-art methods on OpenImages V7~\cite{kuznetsova2020openimagesv7} and RORD~\cite{sagong2022rord} dataset. The best results are in \textbf{bold}.}
    \label{tab:appendix_opiv7_rord_results}
    \resizebox{\linewidth}{!}{
    \begin{tabular}{lccccccccc}
    \toprule
    \multirow{2}{*}{\textbf{Method}} 
        & \multicolumn{4}{c}{\textbf{OpenImages V7}} 
        & \multicolumn{4}{c}{\textbf{RORD}} \\
    \cmidrule(lr){2-5} \cmidrule(lr){6-9}
      & \textbf{BG Sim.} & \textbf{FG Sim.} & \textbf{GPT-Success} & \textbf{GPT-Score}
      & \textbf{BG Sim.} & \textbf{FG Sim.} & \textbf{GPT-Success} & \textbf{GPT-Score} \\
    \midrule
    \rowcolor{gray!15}
    \multicolumn{9}{l}{\textbf{\textit{Dataset-Free Approaches}}} \\
    \midrule
    SD3.5-M~\cite{esser2024dit} & 0.605 & 0.286 & 12.7\% & 16.9 & 0.582 & 0.319 & 3.80\% & 6.87 \\
    + AttentiveEraser {\tiny\color{gray}AAAI'25}~\cite{sun2025attentiveeraser} & 0.559 & 0.276 & 10.5\% & 31.6 & 0.541 & 0.302 & 2.04\% & 23.0 \\
    + DesignEdit {\tiny\color{gray}AAAI'25}~\cite{jia2025designedit} & 0.600 & 0.255 & 24.8\% & 34.1 & 0.597 & 0.273 & 10.2\% & 27.6 \\
    \addlinespace
    \rowcolor{green!15}
    \textbf{+ EraseLoRA (Ours)} & \textbf{0.743} & \textbf{0.151} & \textbf{71.0\%} & \textbf{61.0} & \textbf{0.779} & \textbf{0.138} & \textbf{81.3\%} & {70.2} \\
    \midrule
    \rowcolor{gray!15}
    \multicolumn{9}{l}{\textbf{\textit{Dataset-Driven Approaches}}} \\
    \midrule
    SDXL-Inpainting~\cite{podell2024sdxl} & 0.677 & 0.212 & 27.1\% & 30.9 & 0.645 & 0.234 & 3.83\% & 12.9 \\
    FLUX.1-Fill-dev~\cite{flux2024} & 0.661 & 0.255 & 30.3\% & 33.5 & 0.688 & 0.232 & 9.47\% & 13.8 \\
    PowerPaint {\tiny\color{gray}ECCV'24}~\cite{zhuang2024powerpaint} & 0.669 & 0.217 & 33.2\% & 34.9 & 0.729 & 0.176 & 34.1\% & 37.6 \\
    CLIPAway {\tiny\color{gray}NeurIPS'24}~\cite{ekin2024clipaway} & 0.656 & 0.223 & 33.2\% & 38.4 & 0.744 & 0.156 & 35.8\% & 39.9 \\
    SmartEraser {\tiny\color{gray}CVPR'25}~\cite{jiang2025smarteraser} & 0.709 & 0.185 & 59.5\% & {57.4} & 0.768 & 0.148 & {75.8\%} & \textbf{72.5} \\
    EntityErasure {\tiny\color{gray}CVPR'25}~\cite{zhu2025entityerasure} & 0.679 & 0.204 & 50.5\% & 51.4 & 0.766 & 0.175 & 47.5\% & 49.7 \\
    \bottomrule
    \end{tabular}
    }
\end{table}

\begin{table}[t]
    \centering
    \caption{(Left) Quantitative comparison on paired RemovalBench \cite{wei2025omnieraser} and (Right) inference-time computational cost comparison with previous state-of-the-art methods.}
    \label{tab:appendix_removalbench_inference_cost}
    \begin{minipage}{0.48\linewidth}
        \centering
        \resizebox{!}{2.37cm}{
            \begin{tabular}{lccc}
            \toprule
            \textbf{Method} & \textbf{SSIM ($\uparrow$)} & \textbf{PSNR ($\uparrow$)} & \textbf{LPIPS ($\downarrow$)} \\
            \midrule
            \rowcolor{gray!15}
            \multicolumn{4}{l}{\textbf{\textit{Dataset-Free Approaches}}} \\
            \midrule
            SD3.5-M~\cite{esser2024dit} & 0.772 & 22.3 & 0.185 \\
            + AttentiveEraser~\cite{sun2025attentiveeraser} & 0.780 & 24.5 & 0.181 \\
            + DesignEdit~\cite{jia2025designedit} & 0.782 & 24.9 & 0.168 \\
            \rowcolor{green!15}
            \textbf{+ EraseLoRA (Ours)} & \textbf{0.786} & \textbf{25.1} & \textbf{0.163} \\
            \midrule
            \rowcolor{gray!15}
            \multicolumn{4}{l}{\textbf{\textit{Dataset-Driven Approaches}}} \\
            \midrule
            SDXL-Inpainting~\cite{podell2024sdxl} & 0.726 & 20.7 & 0.430 \\
            FLUX.1-Fill-dev~\cite{flux2024} & 0.757 & 21.6 & 0.212 \\
            PowerPaint~\cite{zhuang2024powerpaint} & 0.751 & 22.9 & 0.213 \\
            CLIPAway~\cite{ekin2024clipaway} & 0.722 & 22.5 & 0.198 \\
            SmartEraser~\cite{jiang2025smarteraser} & 0.744 & 24.2 & 0.168 \\
            OmniEraser~\cite{wei2025omnieraser} & 0.699 & 23.9 & 0.253 \\
            EntityErasure~\cite{zhu2025entityerasure} & 0.723 & 22.4 & 0.208 \\
            \bottomrule
            \end{tabular}
        }
    \end{minipage}
    \hfill
    \begin{minipage}{0.48\linewidth}
        \centering
        \resizebox{!}{2.37cm}{
            \begin{tabular}{lccc}
            \toprule
            \textbf{Method} & \textbf{Params.} & \textbf{VRAM} & \textbf{Latency} \\
            \midrule
            \rowcolor{gray!15}
            \multicolumn{4}{l}{\textbf{\textit{Dataset-Free Approaches}}} \\
            \midrule
            SD3.5-M \cite{esser2024dit} & 2,243 M & 21.9 GB & 4 s \\
            + AttentiveEraser~\cite{sun2025attentiveeraser} & 2,243M & 43.2 GB & 15 s \\
            + DesignEdit \cite{jia2025designedit} & 2,243M & 43.2 GB & 9 s \\
            \rowcolor{green!15}
            + EraseLoRA (Ours) & 2,243 M & 30.9 GB & 4 s \\
            \midrule
            \rowcolor{gray!15}
            \multicolumn{4}{l}{\textbf{\textit{Dataset-Driven Approaches}}} \\
            \midrule
            SDXL-Inpainting \cite{podell2024sdxl} & 2,568 M & 8.3 GB & 5 s \\
            FLUX.1-Fill-dev \cite{flux2024} & 11,902 M & 38.3 GB & 15 s \\
            PowerPaint \cite{zhuang2024powerpaint} & 1,952 M & 4.7GB & 2 s \\
            CLIPAway \cite{ekin2024clipaway} & 1,390 M & 11.3 GB & 2 s \\
            SmartEraser \cite{jiang2025smarteraser} & 1,494 M & 9.7 GB & 2 s \\
            OmniEraser \cite{wei2025omnieraser} & 16,961 M & 35.1 GB & 6 s \\
            EntityErasure \cite{zhu2025entityerasure} & 2,607 M & 13.6 GB & 3 s \\
            \bottomrule
            \end{tabular}
        }
    \end{minipage}
\end{table}

\begin{table}[t]
    \centering
    \caption{EraseLoRA on different diffusion backbones with TTA cost.}
    \label{tab:appendix_robustness_backbones}
    \resizebox{0.6\linewidth}{!}{
    \begin{tabular}{lccccc}
    \toprule
    \multirow{2}{*}{\textbf{Method}} 
    & \multicolumn{2}{c}{\textbf{Metrics}} 
    & \multicolumn{3}{c}{\textbf{TTA Cost}} \\
    \cmidrule(lr){2-3} \cmidrule(lr){4-6}
    & \textbf{BG Sim.($\uparrow$)} 
    & \textbf{FG Sim.($\downarrow$)}
    & \textbf{Param.} 
    & \textbf{VRAM} 
    & \textbf{Opt. Time} \\
    \midrule
    SD1.5~\cite{rombach2022sd} 
    & 0.596 & 0.271 & 858 M & 3.39 GB & - \\
    \rowcolor{green!15}
    + EraseLoRA 
    & \textbf{0.702} & \textbf{0.176} & +6.38 M & +0.68 GB & 117 s \\
    \midrule
    SDXL~\cite{podell2024sdxl} 
    & 0.608 & 0.297 & 2,573 M & 11.5 GB & - \\
    \rowcolor{green!15}
    + EraseLoRA 
    & \textbf{0.730} & \textbf{0.186} & +46.5 M & +5.86 GB & 250 s \\
    \midrule
    SD3.5-M~\cite{esser2024dit} 
    & 0.605 & 0.286 & 2,243 M & 21.9 GB & - \\
    \rowcolor{green!15}
    + EraseLoRA 
    & \textbf{0.743} & \textbf{0.151} & +23.9 M & +9.0 GB & 191 s \\
    \midrule
    FLUX.1~\cite{flux2024} 
    & 0.588 & 0.205 & 11,902 M & 38.5 GB & - \\
    \rowcolor{green!15}
    + EraseLoRA 
    & \textbf{0.760} & \textbf{0.146} & +52.3 M & +20.9 GB & 495 s \\
    \bottomrule
    \end{tabular}
    }
\end{table}

\begin{figure}[t]
    \centering
    \includegraphics[width=1.0\linewidth]{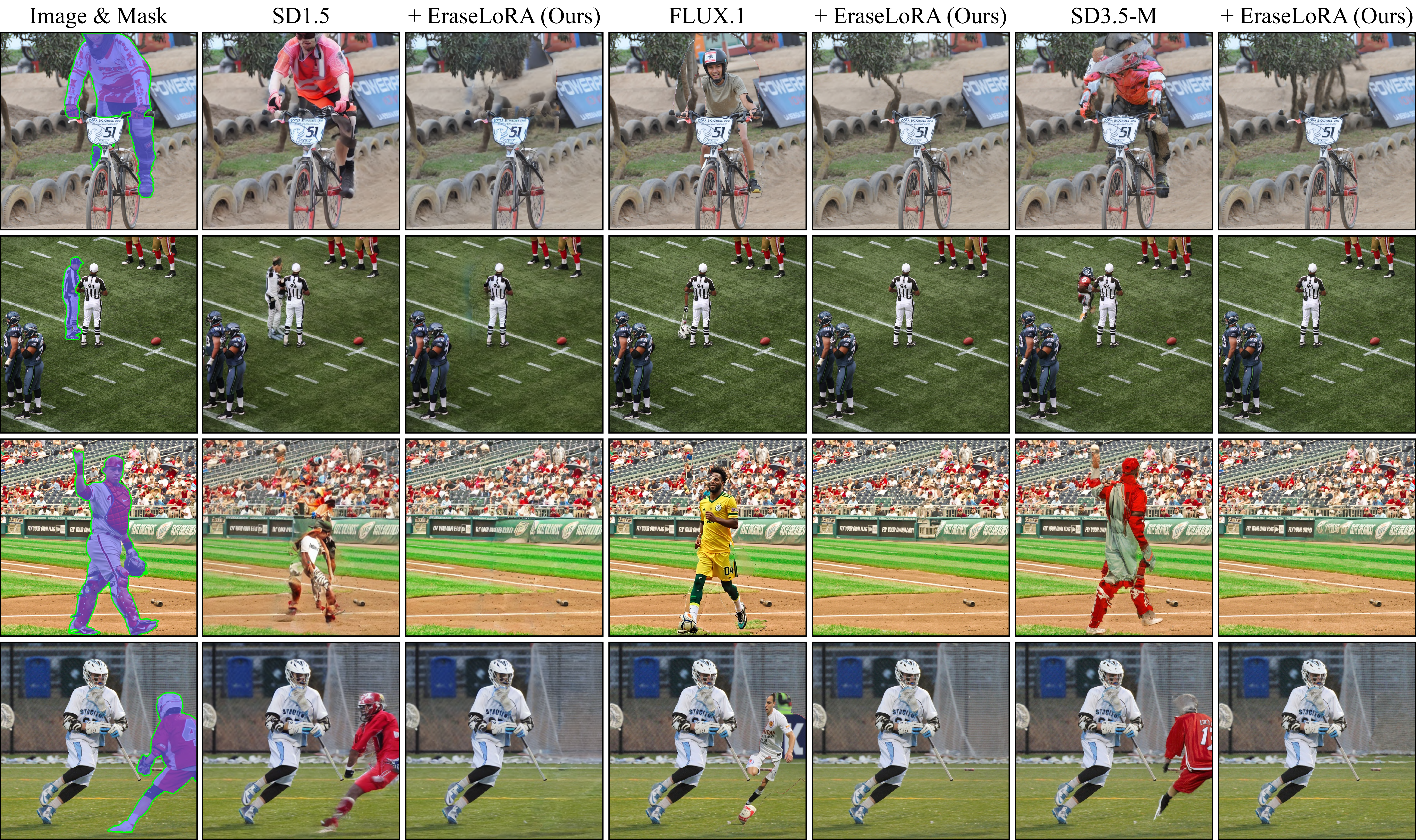} 
    \caption{\textbf{Qualitative results across diffusion architectures.} Consistent clean background restoration on SD1.5, FLUX.1, and SD3.5-M~\cite{esser2024dit, flux2024, rombach2022sd}.}
    \label{fig:appendix_robustness_backbones}
\end{figure}

\textbf{Inference efficiency.}
We compare computational efficiency during inference across recent object removal methods \cite{podell2024sdxl, flux2024, zhuang2024powerpaint, ekin2024clipaway, jiang2025smarteraser, wei2025omnieraser, zhu2025entityerasure, jia2025designedit, sun2025attentiveeraser}. Although EraseLoRA requires additional computation costs during BRSA (\cref{sec:brsa}) due to LoRA adapters (see Tab.~\ref{tab:appendix_robustness_backbones}), EraseLoRA incurs no extra cost at inference by merging LoRA weights into the frozen diffusion backbone's weights \cite{hu2022lora}. (see Tab. \ref{tab:appendix_removalbench_inference_cost}, right). 
While dataset-driven models may offer lower inference cost, EraseLoRA achieves comparable or lower overhead while delivering substantially higher foreground suppression and background fidelity. Therefore, even when a computational gap exists, the quality gains make the trade-off clearly advantageous.

\section{Details of Ablation Study} \label{sec:appendix_ablation}
\subsection{Flexibility} \label{subsec:appendix_flexibility}
EraseLoRA is designed as a model-agnostic framework that can be plugged into different components of the object-removal pipeline. In the following, we examine its flexibility by varying (i) the underlying diffusion backbone, (ii) the MLLM used for background-aware reasoning, and (iii) the Tag2Mask pipeline, and show that EraseLoRA yields consistent improvements across these choices.

\textbf{Diffusion architectures.}
We evaluate EraseLoRA on four representative text-to-image diffusion backbones, including SD1.5~\cite{rombach2022sd}, SDXL~\cite{podell2024sdxl}, SD3.5-M~\cite{esser2024dit}, and FLUX.1~\cite{flux2024}, thereby demonstrating that EraseLoRA robustly reconstructs foreground-free background across diverse architectures. EraseLoRA consistently improves the performance, showing at least a 17.8\% increase in BG similarity and a 28.8\% decrease in FG similarity, across all employed backbones (see Tab. \ref{tab:appendix_robustness_backbones}). Notably, this improvement is more pronounced in text-to-image diffusion backbones with superior text-to-image alignment power (\emph{e.g.}, SD3.5-M~\cite{esser2024dit} and FLUX.1~\cite{flux2024}), as they provide more precise guidance for identifying and reconstructing background.
This backbone-agnostic behavior is also clearly observed in qualitative results, where it stably removes target objects without foreground traces or noise, while preserving fine details and global background coherence (see Fig. \ref{fig:appendix_robustness_backbones}).

\begin{figure}[t]
    \centering
    \includegraphics[width=\linewidth]{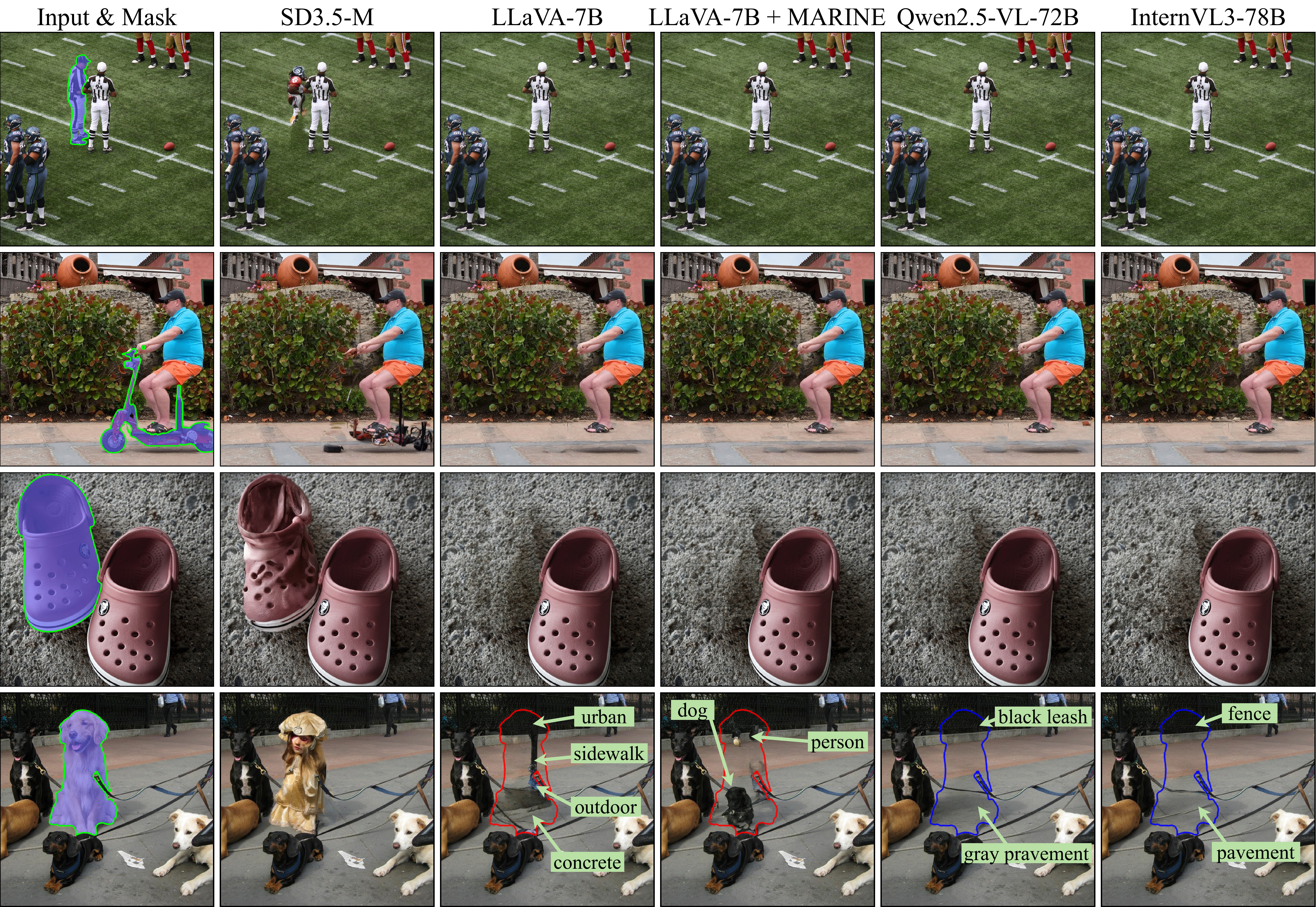} 
    \caption{\textbf{Qualitative results across MLLMs.} Clean background reconstruction and strong foreground suppression are consistently achieved across diverse MLLMs \cite{liu2023llava, zhao2025marine, bai2025qwen2.5vl, zhu2025internvl3}. MLLMs with strong background-aware reasoning~\cite{bai2025qwen2.5vl, zhu2025internvl3} exhibit superior removal quality (blue overlays) by providing accurate background cues (green boxes), whereas smaller MLLMs~\cite{liu2023llava, zhao2025marine} often fail to remove the object (red overlays) due to inaccurate background cues (\emph{e.g.}, dog or person).}
    \label{fig:appendix_robustness_mllm}
\end{figure}

\textbf{MLLMs.}
EraseLoRA yields noticeable improvements in different MLLMs \cite{liu2023llava, zhao2025marine, bai2025qwen2.5vl, zhu2025internvl3}, even when using lightweight models (see Tab. \ref{tab:ab_robustness_mllms_tag2mask}). This tendency is also qualitatively confirmed, where EraseLoRA suppresses object generation and restores the background coherently guided by MLLM-driven background cues (see Fig. \ref{fig:appendix_robustness_mllm}, left).
While we test MARINE~\cite{zhao2025marine}, a hallucination-mitigated model, to examine the impact on removal quality, we observed that background-aware reasoning power, the ability to accurately identify and classify background subtypes among diverse candidates, is a more critical factor. Large MLLMs (\emph{e.g.}, Qwen2.5-VL-72B~\cite{bai2025qwen2.5vl} and InternVL3-78B~\cite{zhu2025internvl3}) successfully remove the target object and reconstruct background based on precisely inferred background subtypes, unlike MARINE~\cite{zhao2025marine} which fails to remove objects effectively due to inaccurate background cues (see Fig. \ref{fig:appendix_robustness_mllm}, last row).

\begin{figure}[t]
    \centering
    \includegraphics[width=1.00\linewidth]{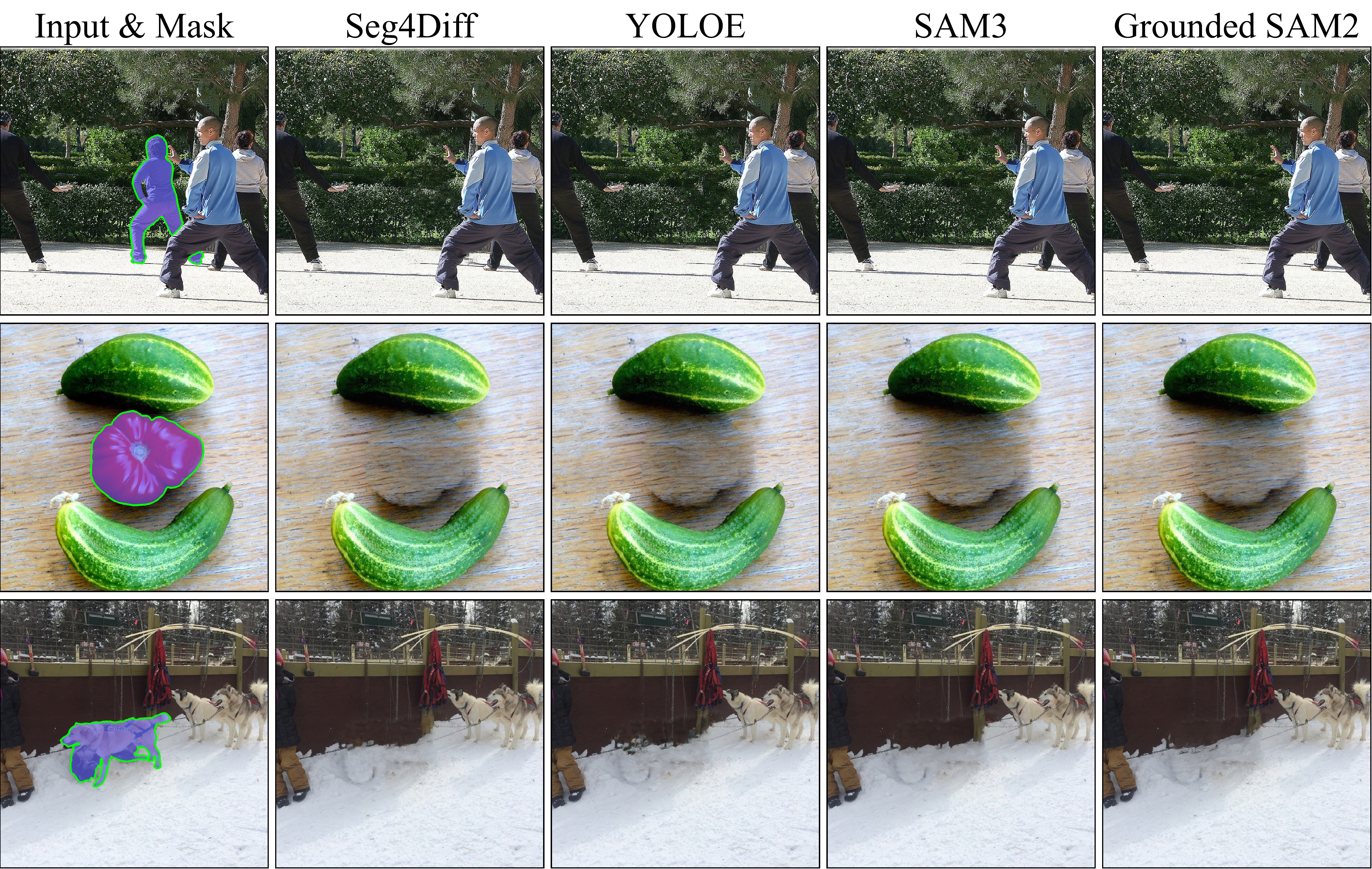} 
    \caption{\textbf{Qualitative results across Tag2Mask models.} Different Tag2Mask models~\cite{kim2025seg4diff, wang2025yoloe, liu2024groundingdino, ravi2025sam2, carion2025sam3} reliably localize non-target foreground regions, enabling complete background reconstruction without foreground traces.}
    \label{fig:appendix_tag2mask}
\end{figure}

\textbf{Tag2Mask models.}
We further validate that the proposed framework remains effective with different Tag2Mask models in BFE (\cref{sec:bfe}), including Seg4Diff~\cite{kim2025seg4diff}, YOLOE~\cite{wang2025yoloe}, SAM3~\cite{carion2025sam3} and Grounded SAM2 (Grounding DINO \cite{liu2024groundingdino} and SAM2~\cite{ravi2025sam2}). 
Across all Tag2Mask variants, EraseLoRA consistently improves background reconstruction and foreground suppression, yielding at least 10.0\% gains in BG Sim. and 28.3\% reductions in FG Sim. over the SD3.5-M~\cite{esser2024dit} baseline (see Tab.~\ref{tab:ab_robustness_mllms_tag2mask}, right).
Notably, Grounded SAM2 achieves the best performance, improving BG Sim. by up to 22.8\% and reducing FG Sim. by up to 47.2\%, resulting in the cleanest and most faithful background reconstruction (see Fig.~\ref{fig:appendix_tag2mask}).

These results show that EraseLoRA is model-agnostic, supporting plug-and-play object removal without depending on specific external modules.

\begin{figure}[t]
    \centering
    \includegraphics[width=\linewidth]{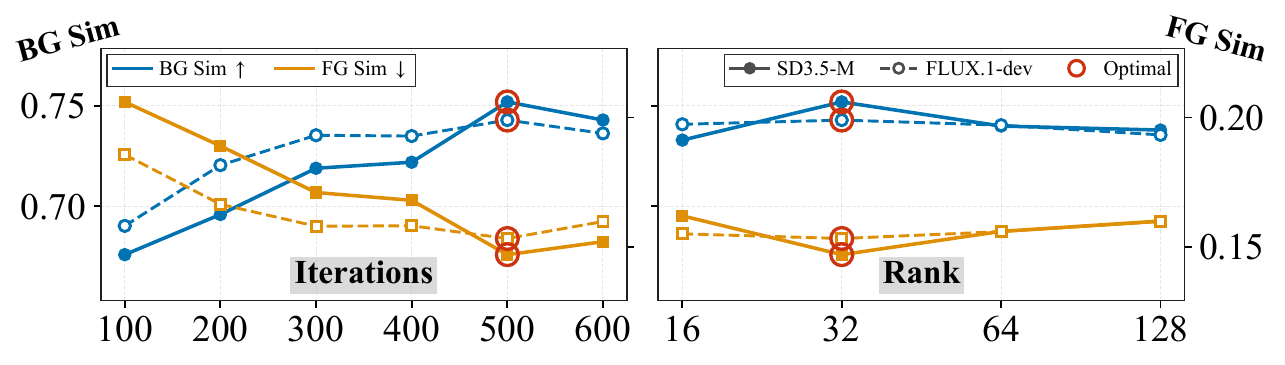} 
    \caption{\textbf{Effect of test-time optimization \cite{wang2021tta} capacity.} Varying LoRA \cite{hu2022lora} rank and number of iterations shows that 500 iterations and LoRA rank 32 achieve the best performance on SD3.5-M \cite{esser2024dit} and FLUX.1 \cite{flux2024}.}
    \label{fig:appendix_tta_param}
\end{figure}

\subsection{Adaptation Capacity}
We vary two key factors in test-time optimization (BRSA; \cref{sec:brsa}) on SD3.5-M~\cite{esser2024dit} and FLUX.1~\cite{flux2024}: (1) the LoRA rank \cite{hu2022lora}, which controls the learnable capacity of adapters, and (2) the number of test-time adaptation \cite{wang2021tta} iterations, which determines how long the model adapts to background cues.

\textbf{LoRA rank.}
From experiments with ranks \{16, 32, 64, 128\}, rank 32 yields the best trade-off, achieving the strongest foreground suppression and the most consistent background reconstruction (see Fig. \ref{fig:appendix_tta_param}). Larger ranks such as 64 or 128 offer no meaningful gains while incurring higher optimization cost, so we adopt rank 32 as the default configuration.

\textbf{TTA iterations.}
Although longer optimization generally improves reconstruction performance, the marginal gains diminish relative to the additional time cost. Hence, we adopt 500 iterations as a practical balance between quality and efficiency (see Fig. \ref{fig:appendix_tta_param}).

\subsection{Discussion} \label{subsec:appendix_discussion}
\textbf{Multiple objects removal.}
When the mask contains multiple objects to erase, EraseLoRA removes all targets jointly and reconstructs each region with coherent background subtypes. Because the adaptation operates per background rather than per instance, its performance remains stable regardless of the type or number of masked objects, requiring no modification to the framework (see Fig. \ref{fig:appendix_video} {(b)}).

\begin{figure}[t]
    \centering
    \includegraphics[width=\linewidth]{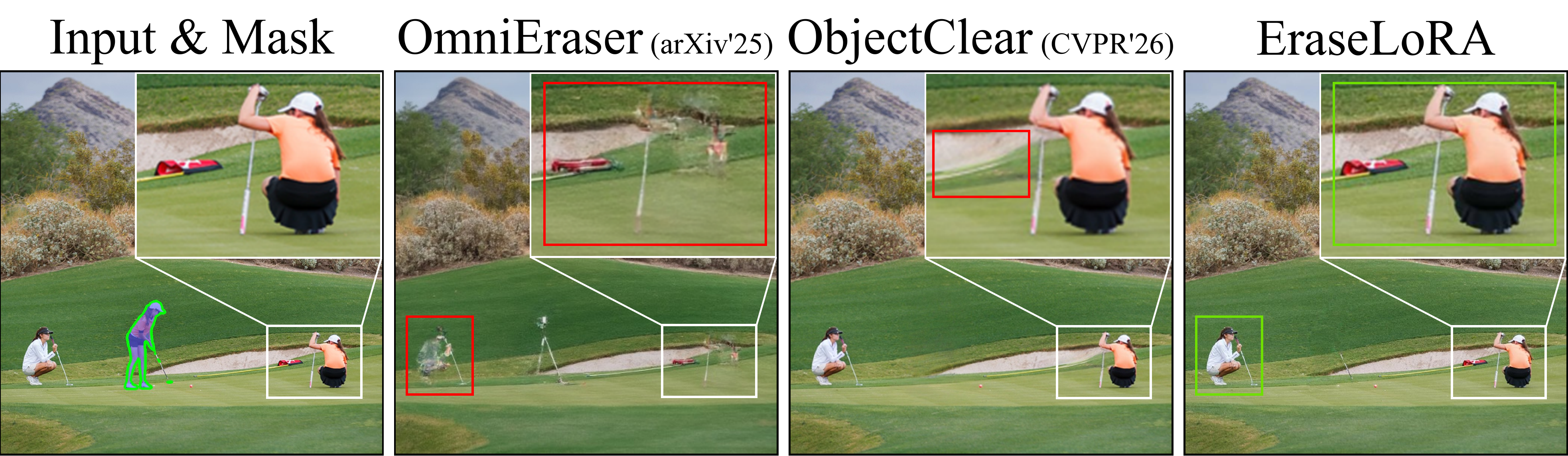}
    \caption{\textbf{Trade-off in effect-aware removal.} Effect-aware methods~\cite{wei2025omnieraser, zhao2026objectclear} can target object-induced effects, but often distort target-unrelated regions (red), whereas EraseLoRA preserves them (green).}
    \label{fig:appendix_effectaware_tradeoff}
\end{figure}

\textbf{Trade-off in effect-aware removal.}
Effect-aware methods~\cite{wei2025omnieraser, zhao2026objectclear} explicitly target object-induced effects such as shadows or reflections, but may distort target-unrelated regions (see Fig.~\ref{fig:appendix_effectaware_tradeoff}). In contrast, EraseLoRA prioritizes preserving such regions by editing only the masked area. When object-induced effects should also be removed, they can be handled by expanding the mask or using interactive control (see Fig.~\ref{fig:appendix_human_points}).

\textbf{Misclassification of background subtypes.} While our method utilizes MLLM's background-aware reasoning capabilities, the removal quality may be affected if foreground and background tags are misclassified. In such cases, EraseLoRA follows incorrect cues and regenerates residual object traces. This issue can be alleviated by employing larger MLLMs \cite{bai2025qwen2.5vl, zhu2025internvl3} with stronger background-aware reasoning or through interactive control (see last row of Fig.~\ref{fig:appendix_robustness_mllm} and Fig.~\ref{fig:appendix_human_points}).

\section{Limitations and Future Works} \label{sec:limitations_and_futureworks}
\subsection{Limitations} \label{subsec:limitations}

\begin{figure}[t]
    \centering
    \includegraphics[width=1.0\linewidth]{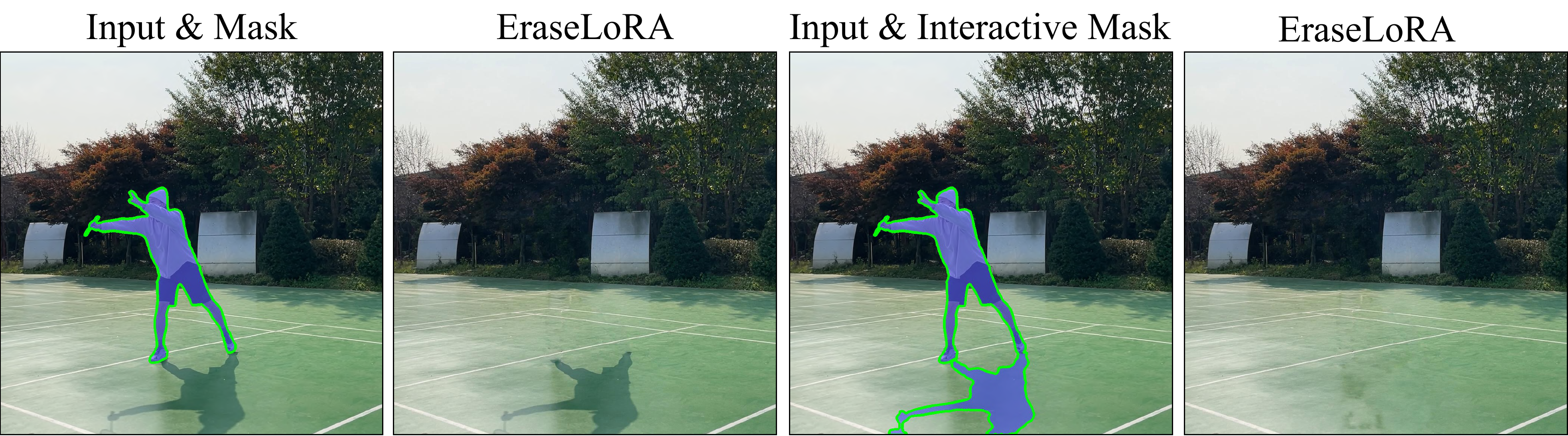} 
    \caption{\textbf{Failure case and practical solution.} While physical effects (\emph{e.g.}, shadows) outside the initial mask can leave subtle traces (left), using an interactive mask that encompasses these effects ensures complete object removal (right).}
    \label{fig:appendix_limitations_failure}
\end{figure}

\begin{figure}[t]
    \centering
    \includegraphics[width=\linewidth]{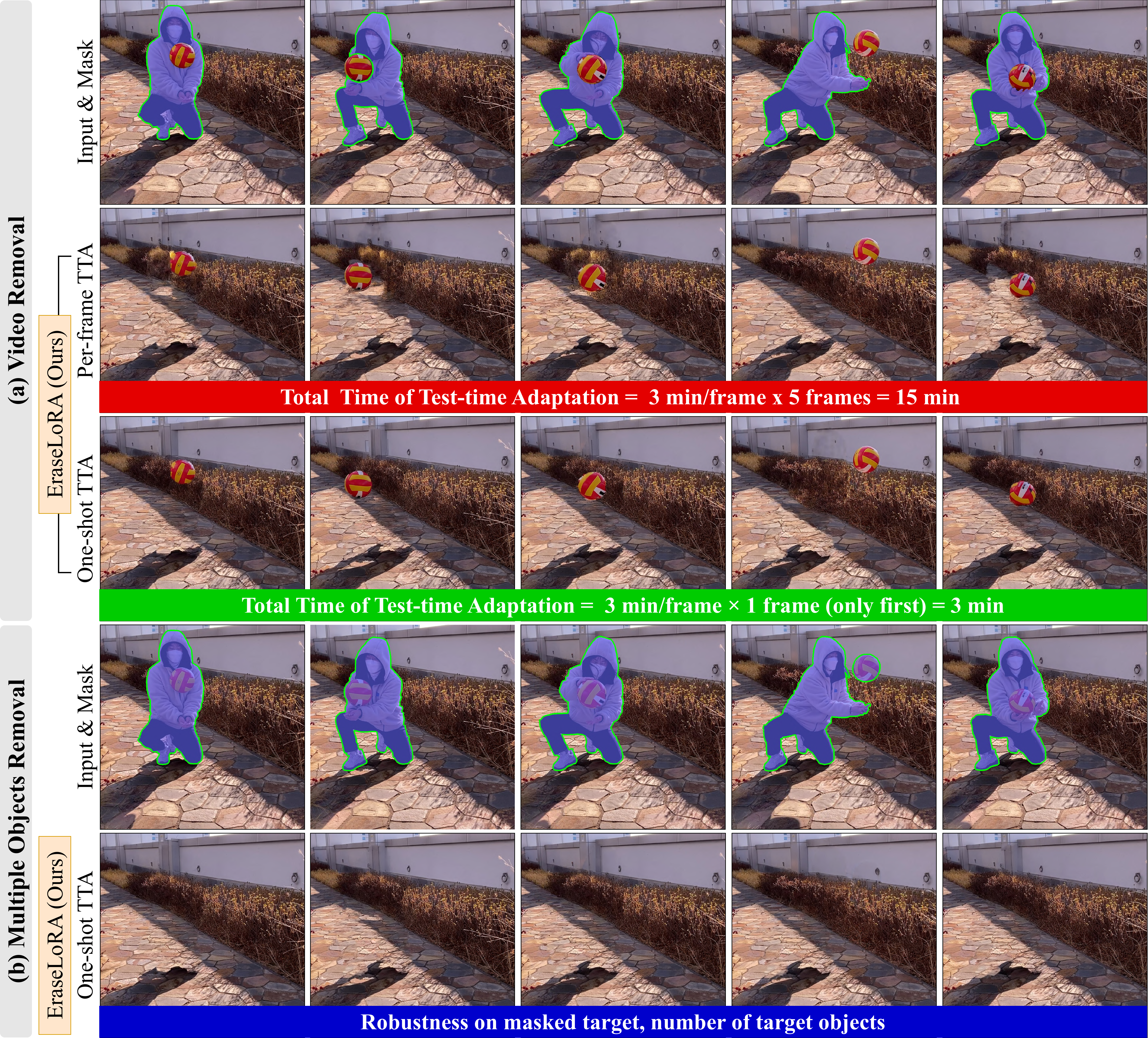} 
    \caption{\textbf{Efficiency of video extension.} For video frames sharing similar background context, one-shot optimization on a single frame can be reused across the sequence, achieving performance comparable to per-frame optimization while reducing adaptation cost by the number of frames. Moreover, EraseLoRA remains robust for multiple object removal without additional optimization.}
    \label{fig:appendix_video}
\end{figure}

\textbf{Presence of object effects.} EraseLoRA removes the target object and synthesizes plausible background texture, but does not explicitly handle physical effects caused by the object outside the mask (\emph{e.g.}, shadows, lighting distortion, reflections). 
Thus, subtle traces may remain if these effects are not included in the mask. A practical solution is to use an interactive mask that encompasses both the object and its effects, which helps achieve more seamless removal (see Fig.~\ref{fig:appendix_limitations_failure}).

\textbf{Additional computational overhead.}
EraseLoRA requires background-aware reasoning from MLLMs \cite{liu2023llava, zhu2025internvl3, bai2025qwen2.5vl} and test-time optimization \cite{wang2021tta} with LoRA adapters \cite{hu2022lora}, introducing extra computation compared to training-free object removal methods \cite{esser2024dit, jia2025designedit, sun2025attentiveeraser}. Since the optimization is performed per background context, the cost scales with the number of different backgrounds encountered.
However, this overhead can be effectively mitigated through several efficiency-oriented strategies. By selectively utilizing MLLMs of various scales, such as using LLaVA-7B \cite{liu2023llava} instead of the default InternVL3-78B \cite{zhu2025internvl3}, the parameter scale is reduced by approximately 11$\times$ while maintaining over 98.0\% and 91.4\% of the performance in BG Sim. and FG Sim., respectively (see Tab. \ref{tab:ab_robustness_mllms_tag2mask}, left).

Furthermore, we adopt an early stopping (E.S.) rule based on a standard elbow-based criterion to determine the stopping point automatically for computational efficiency, as the optimization exhibits a clear diminishing-return pattern.
For each sample, we track the optimization objective $\mathcal{L}_{\text{total}}$ and identify the elbow of its smoothed trajectory, which marks the transition from rapid improvement to a near-plateau regime. Importantly, this criterion is not tailored to our method; rather, it is a conventional curve-based stopping rule applied directly to the loss dynamics. Using this criterion, the average optimization length is reduced from 500 to approximately 140 iterations, while preserving more than 96.5\% of the final BG Sim. relative to the full optimization budget. While E.S. rule offers a favorable efficiency-performance trade-off, we do not adopt it in the main experiments because our primary goal in this paper is to maximize final restoration quality under a fixed optimization budget. In practice, early stopping introduces a small but non-negligible performance drop compared with the full 500-iteration optimization. Nevertheless, we believe it provides a practical option for follow-up studies and for deployment scenarios where computational overhead is a more critical concern.

\subsection{Future Works} \label{subsec:future_works}
% \textbf{Object removal in video.}
Although EraseLoRA incurs additional computational overhead, its optimization is performed per background rather than per image. 
In video sequences where many frames share similar backgrounds, this allows the optimization cost to be reused across frames, making video object removal a promising next step.

To validate this potential, we apply test-time optimization \cite{wang2021tta} only to a single representative frame and reuse the adapted model for the remaining frames. As shown in Fig. \ref{fig:appendix_video}, the outputs remain comparable to per-frame optimization, where the model is independently adapted for every frame. This demonstrates that leveraging shared background context makes video object removal an efficient extension of EraseLoRA (see Fig.~\ref{fig:appendix_video} {(a)}).

\section{Additional qualitative results} \label{sec:appendix_qualitative_results}
This section provides extended qualitative comparisons between EraseLoRA and various baseline methods~\cite{podell2024sdxl, flux2024, zhuang2024powerpaint, ekin2024clipaway, jiang2025smarteraser, wei2025omnieraser, zhu2025entityerasure,esser2024dit, jia2025designedit, sun2025attentiveeraser}. All results were generated using the same experimental setup, including the baselines and benchmarks~\cite{kuznetsova2020openimagesv7, sagong2022rord, wei2025omnieraser} detailed in \cref{subsec:appendix_experiment_setup}.

\begin{figure}[p]
    \centering
    \includegraphics[height=0.9\textheight, keepaspectratio]{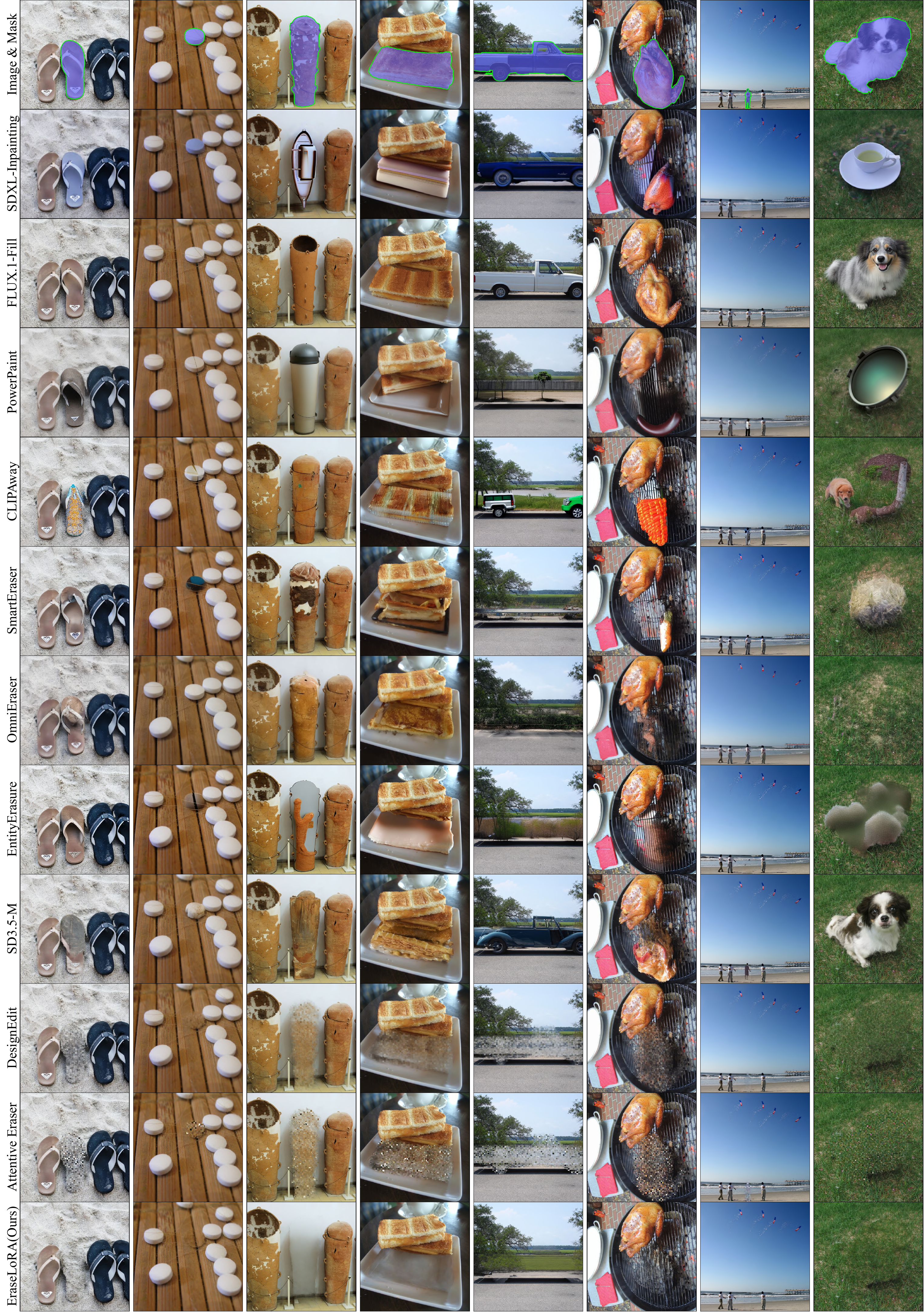} 
    \caption{Additional qualitative comparison with dataset-driven and dataset-free methods \cite{podell2024sdxl, flux2024, zhuang2024powerpaint, ekin2024clipaway, jiang2025smarteraser, wei2025omnieraser, zhu2025entityerasure, esser2024dit, jia2025designedit, sun2025attentiveeraser} on OpenImages V7 \cite{kuznetsova2020openimagesv7} dataset.}
    \label{fig:appendix_qualitative_results1}
\end{figure}

\begin{figure}[p]
    \centering
    \includegraphics[height=0.9\textheight, keepaspectratio]{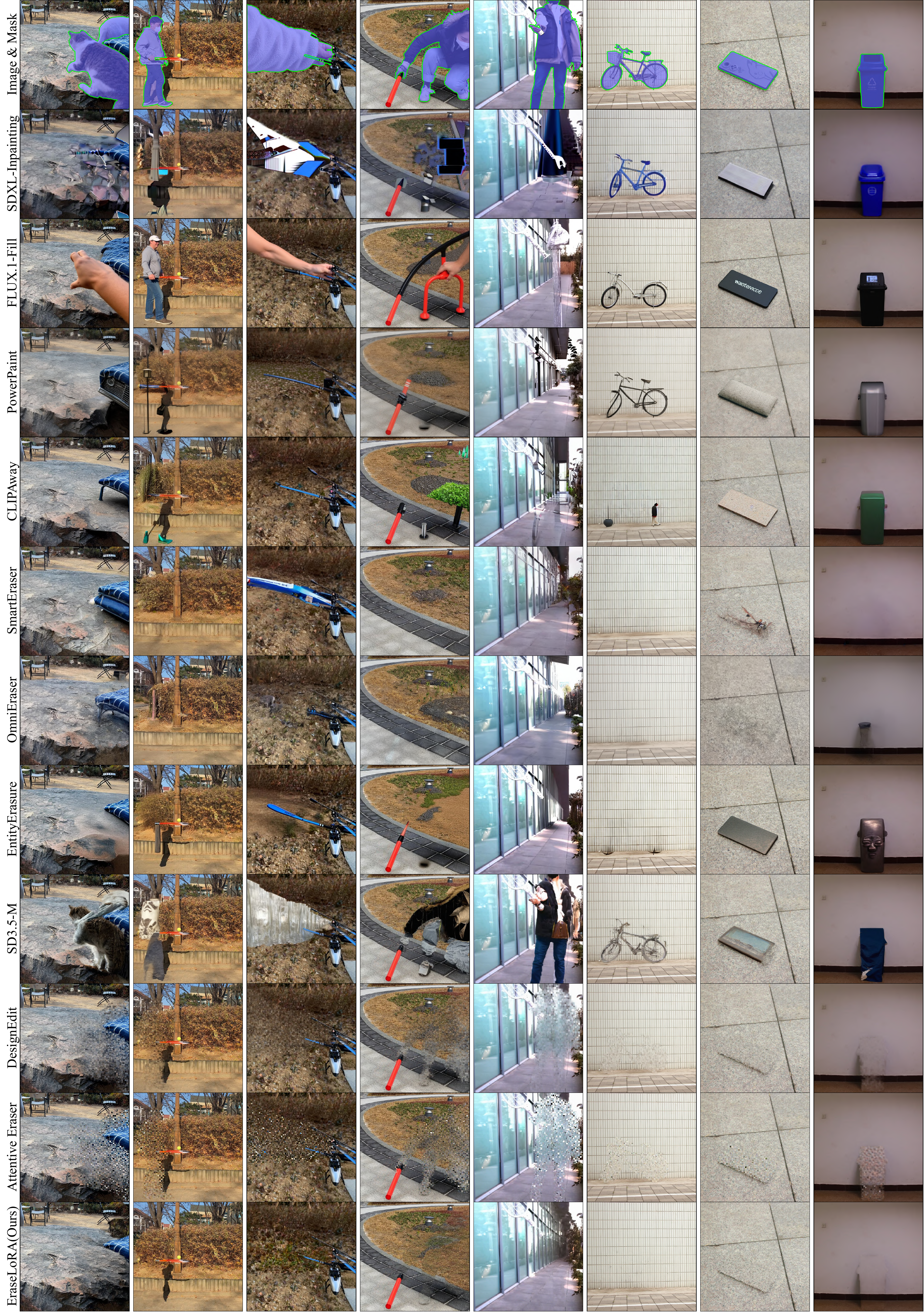} 
    \caption{Additional qualitative comparison with dataset-driven and dataset-free methods \cite{podell2024sdxl, flux2024, zhuang2024powerpaint, ekin2024clipaway, jiang2025smarteraser, wei2025omnieraser, zhu2025entityerasure, esser2024dit, jia2025designedit, sun2025attentiveeraser} on RORD \cite{sagong2022rord} and RemovalBench \cite{wei2025omnieraser} datasets.}
    \label{fig:appendix_qualitative_results2}
\end{figure}

EraseLoRA clearly removes target objects without leaving semantic traces and reconstructs the background with artifact-free textures.
In contrast, previous dataset-free methods \cite{esser2024dit, jia2025designedit, sun2025attentiveeraser} tend to hallucinate foreground-like patterns or overly smooth textures. 
Visual comparisons are shown in Fig.~\ref{fig:appendix_qualitative_results1} and Fig.~\ref{fig:appendix_qualitative_results2}.

\end{document}